\documentclass[10pt,journal,compsoc]{IEEEtran}
\usepackage[nocompress]{cite}
\usepackage{lineno}
\usepackage{amsmath}
\usepackage{amssymb}
\usepackage{commath}
\usepackage{tabularx}
\usepackage{booktabs}
\usepackage{float}
\usepackage{algorithm}
\usepackage{algpseudocode}
\usepackage{graphicx}
\usepackage{subfigure}
\usepackage{color}
\usepackage{CJK}
\usepackage{cite}
\usepackage{multirow}
\usepackage{multicol}
\usepackage{soul}
\usepackage{tablefootnote}

\usepackage[pagebackref=false,breaklinks=false,letterpaper=true,colorlinks,bookmarks=false]{hyperref}

\begin{document}
\title{3D Registration in 30 Years: A Survey\IEEEcompsocitemizethanks{\IEEEcompsocthanksitem This work is supported in part by National Natural Science Foundation of China (No. 62372377). \IEEEcompsocthanksitem Jiaqi Yang, Chu'ai Zhang, Zhengbao Wang, Xinyue Cao, Xuan Ouyang, Xiyu Zhang, Zhenxuan Zeng, Zhiyi Xia and Yanning Zhang are with the School of Computer Science, Northwestern Polytechnical University, China. E-mail: \{cazhang, npu-wzb, caoxinyue, ouyangxuan, 2426988253, zengzhenxuan, xiazhiyi \}@mail.nwpu.edu.cn; \{jqyang, ynzhang\}@nwpu.edu.cn. 
\IEEEcompsocthanksitem Zhao Zeng and Borui Lu are with the School of Electronics and Control Engineering, Chang’an University, China. E-mail: \{zhaozeng, 2024232056\}@chd.edu.cn.
\IEEEcompsocthanksitem Qian Zhang is with School of Architecture \& Urban Planning, Shenzhen University, China. E-mail: \{hangfanzq\}@163.com. (\textit{Corresponding author: Qian Zhang})
\IEEEcompsocthanksitem Yulan Guo is with the School of Electronics and Communication
Engineering, Sun Yat-Sen University, China. E-mail: \{guoyulan\}@sysu.edu.cn.

}}

\author{Jiaqi Yang, Chu'ai Zhang, Zhengbao Wang, Xinyue Cao, Xuan Ouyang, Xiyu Zhang, Zhenxuan Zeng, Zhao Zeng, Borui Lu, Zhiyi Xia, Qian Zhang, Yulan Guo,~\IEEEmembership{Senior Member,~IEEE}, and Yanning Zhang,~\IEEEmembership{Fellow,~IEEE}}


\markboth{Journal of \LaTeX\ Class Files,~Vol.~14, No.~8, August~2015}%
{Shell \MakeLowercase{\textit{et al.}}: Bare Demo of IEEEtran.cls for Computer Society Journals}
\IEEEtitleabstractindextext{%
\begin{abstract}
 3D point cloud registration is a fundamental problem in computer vision, computer graphics, robotics, remote sensing, and etc. Over the last thirty years, we have witnessed the amazing advancement in this area with numerous kinds of solutions. Although a handful of relevant surveys have been conducted, their coverage is still limited. In this work, we present a comprehensive survey on 3D point cloud registration, covering a set of sub-areas such as pairwise coarse registration, pairwise fine registration, multi-view registration, cross-scale registration, and multi-instance registration. The datasets, evaluation metrics, method taxonomy, discussions of the merits and demerits, insightful thoughts of future directions are comprehensively presented in this survey. The regularly updated project page of the survey is available at \url{https://github.com/Amyyyy11/3D-Registration-in-30-Years-A-Survey}.
\end{abstract}
}
\maketitle
\IEEEdisplaynontitleabstractindextext
\IEEEpeerreviewmaketitle
\IEEEraisesectionheading{\section{Introduction}\label{sec:intr}}
\IEEEPARstart{A}{ligning} 3D point clouds to a unified coordinate system, known as 3D point cloud registration, is a fundamental problem in numerous areas such as computer vision, computer graphics, robotics and remote sensing.  Aligned point clouds offer two key results: 1) a more complete point cloud for reconstruction, information fusion, and error measurement; 2) a six-degree-of-freedom (6-DoF) pose for robust pose estimation, 3D tracking, object/place localization, and motion-flow estimation. With the development of 3D active and passive acquisition technology (e.g., Intel's RealSense, Apple's iPhone series), 3D point cloud registration has attracted an increasing research attention on this topic during the last three decades.

\begin{figure}
\centering
    \includegraphics[width=0.38\textwidth]{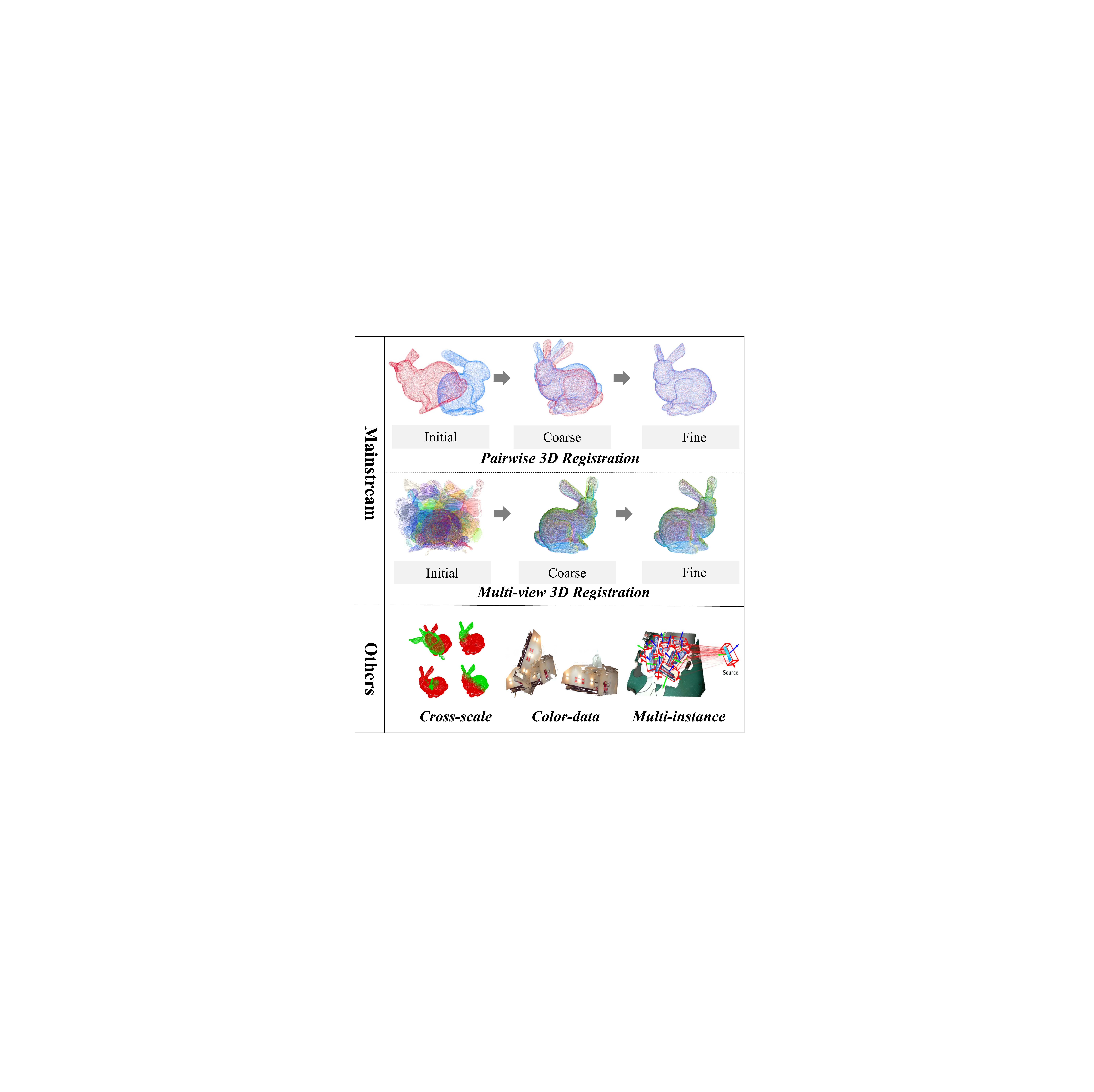}
    \caption{Typical 3D registration problems.}
    \label{fig:Typical tasks}
\end{figure}

In particular, there are several sub-branches towards robust 3D point cloud registration, depending on either data acquisition or application scenarios (Fig.~\ref{fig:Typical tasks}). From the perspective of handled data sequences, pairwise registration focuses on aligning two point clouds, while multi-view registration aligns more than two sequential or unordered multiple point clouds. From the perspective of error minimization, coarse registration roughly aligns point clouds with relatively large pose variation, while fine registration usually focuses on minimizing small residual errors. From the perspective of methodologies, early methods design hand-crafted optimization or heuristic methods, while recent methods resort to deep learning approaches. There are also other perspectives for investigating the registration problem, such as feature learning, correspondence learning, and robust 6-DoF pose estimation. Therefore, there are many methods and research topics in the 3D point cloud registration realm.

Existing surveys are focused on either different parts or a limited scope of point cloud registration tasks. For instance, an early review~\cite{huang2021comprehensive} covers various aspects of point cloud registration but lacks a thorough analysis of the interconnections between subfields, failing to systematically reveal the intrinsic relationships and interactions among them. A recent review~\cite{lyu2024rigid} recaps commonly used datasets and evaluation metrics but lacks performance comparisons in unified experimental settings, failing to demonstrate the advantages and limitations of different methods under consistent conditions. As such, they fail to cover the literature from the last three decades from a more comprehensive perspective. 

\begin{figure*}[t]
    \includegraphics[width=\textwidth]{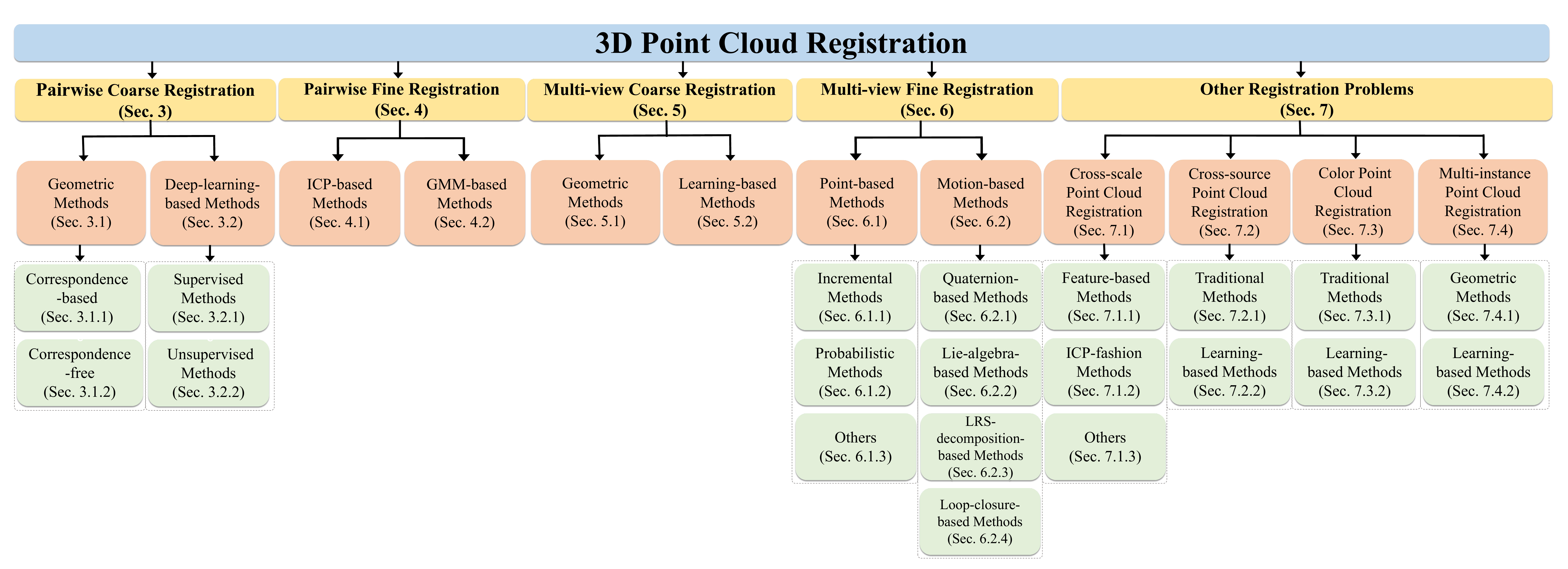}
    \caption{A taxonomy of 3D point cloud registration methods.}
    \label{fig:Taxonomy 3D PCR }
\end{figure*}

To fill the gap, we present a comprehensive survey on 3D registration methods in the last decades. The major contributions are summarized as follows.
\begin{itemize}
\item {\textbf {Thorough review and new taxonomy.}} To the best of our knowledge,  as shown in Fig.~\ref{fig:Taxonomy 3D PCR }, this is the first survey paper to comprehensively review point cloud registration methods, covering a set of subareas such as pairwise coarse registration, pairwise fine registration, multi-view registration, cross-scale registration, and multi-instance registration. It offers a systematic taxonomy and a broad literature coverage.
\item {\textbf {Benchmark overview and performance comparison.}} Popular benchmark datasets and performance evaluation metrics for point cloud registration are systematically summarized. A set of comparative results of representative state-of-the-art methods on standard benchmarks are also reported.
\item {\textbf {Outlook on future directions.}} The traits, merits, and demerits of the existing methods have been highlighted. We also present insightful discussions on current challenges and several future research directions to inspire follow-up works in this field. 
\end{itemize}

The remainder of the paper is organized as follows. Sec.~\ref{sec:Background} reviews point cloud registration datasets and evaluation metrics. Sec.~\ref{sec:Pairwise Coarse Registration} introduces pairwise coarse registration methods, including correspondence-based and correspondence-free approaches. Sec.~\ref{sec:Pairwise Fine Registration} discusses pairwise fine registration methods, focusing on ICP-based and GMM-based methods. Sec.~\ref{sec:Multi-view Coarse Registration} presents multi-view coarse registration methods, covering geometric and deep learning-based approaches. Sec.~\ref{sec:Multi-view Fine Registration} introduces multi-view fine registration methods, including point-based and motion-based methods. Sec.~\ref{sec:Other Registration Problems} introduces other registration problems, such as cross-scale, cross-source, color point cloud, and multi-instance registration. Sec.~\ref{sec:sum} discusses challenges and opportunities in the field. Finally, Sec.~\ref{sec:conc} draws conclusions.
\section{Background}\label{sec:Background}
\subsection{Basic Concepts}
\textbf{Point cloud}. A point cloud $\mathbf{P}$ comprises a set of 3D points $\{\mathbf{p}_i|i=1,2,...,N\}$, where $\mathbf{p}_i \in \mathbb{R}^3$. It is a discrete representation of the 3D continuous surface, which is unorganized and permutation-insensitive.
\\\\\textbf{Point cloud registration}. Given a set of point clouds $\{\mathbf{P}_1,\mathbf{P}_2,...,\mathbf{P}_N\}$ of the same object or scene. The goal of point cloud registration is to determine multiple transformations $\{\mathbf{T}_1, \mathbf{T}_2,...,\mathbf{T}_N\}\in SE(3)$ composed of the corresponding rotation matrices $\{\mathbf{R}_1,\mathbf{R}_2,...,\mathbf{R}_N\} \in SO(3)$ and corresponding translation vectors $\{\mathbf{t}_1,\mathbf{t}_2,...,\mathbf{t}_N\} \in \mathbb{R}^3$: $\{\mathbf{P}_1,\mathbf{P}_2,...,\mathbf{P}_N\} \rightarrow \mathbf{P}_{reg}$, which aligns these point clouds to a unified coordinate system. It generates a more complete point cloud and a set of pose parameters.

\subsection{Datasets}
A large number of datasets have been collected to evaluate the performance of geometric and deep-learning methods for various registration tasks. Depending on the popularity in the community, we list several standard datasets used for the registration problems investigated in this survey in Table~\ref{Tab:Datasets}.

\begin{table*}[t]
\begin{minipage}{\textwidth}
 \centering
  \caption{A summary of representative datasets for 3D registration problems.}
  \renewcommand{\arraystretch}{1.2}
  \renewcommand{\thempfootnote}{\arabic{mpfootnote}}
  \resizebox{\textwidth}{!}{
\begin{tabular}{|c|c|c|c|c|c|c|}
        \hline
        Name & Year & \# Samples & Acquisition & Type & Nuisances & Application scenario \\
        \hline
        Stanford 3D Scanning Repository~\cite{curless1996volumetric} & 1996 & 52 & LiDAR & Object & Real noise & Pairwise fine \& Multi-view \& Cross-scale registration \\
        \hline
        U3M~\cite{mian2006novel} & 2006 & 496 & LiDAR & Object & Limited overlap \& Self-occlusion & Pairwise coarse registration \\
        \hline
        U3OR~\cite{mian2006three,mian2010repeatability} & 2006 & 188 & LiDAR & Indoor & Occlusion \& Clutter & Pairwise coarse registration \\
        \hline
        QuLD~\cite{taati2011local} & 2011 & 240 & LiDAR & Outdoor & Occlusion \& Clutter \& Noise & Pairwise coarse registration \\
        \hline
        RGB-D scenes~\cite{lai2014unsupervised} & 2011 & 300 & Kinect & Indoor & Occlusion \& Real noise & Pairwise fine registration \\
        \hline
        EPFL Statue~\cite{epfl2012statues} & 2012 & 55 & Synthetic & Object & Limited overlap \& Occlusion \& Real noise & Pairwise fine registration \\
        \hline
        RGB-D SLAM~\cite{sturm2012benchmark} & 2012 & 39 & Kinect & Indoor & Occlusion \& Real noise & Pairwise fine registration \\
        \hline
        KITTI~\cite{geiger2012we} & 2012 & 555 & LiDAR & Outdoor & Clutter \& Occlusion \& Real noise & Pairwise coarse \& fine \& Multi-view registration\\
        \hline
        ETH~\cite{pomerleau2012challenging} & 2012 & 713 & LiDAR & Outdoor & Limited overlap \& Clutter \& Occlusion \& Real noise & Pairwise coarse \& fine \& Multi-view registration \\
        \hline
        B3R~\cite{tombari2013performance} & 2013 & 54 & Synthetic &
Object & Gaussian noise \& Mesh decimation & Pairwise coarse registration \\
        \hline
        Space time~\cite{tombari2013performance} & 2013 & 120 & Synthetic & Object & Occlusion & Pairwise coarse registration \\
        \hline
        ModelNet40~\cite{wu20153d} & 2015 & 12311 & Synthetic & Object & Partial missing data \& Simulated noise & Pairwise coarse \& fine \& Multi-instance registration \\
        \hline
        ScanNet~\cite{dai2017scannet} & 2017 & 1513 & Kinect & Indoor & Occlusion \& Real noise & Multi-view registration \\
        \hline
        3DMatch~\cite{zeng20173dmatch} & 2017 & 1623 & Kinect & Indoor & Occlusion \& Real noise & Pairwise coarse \& fine \& Multi-view registration \\
        \hline
         Scan2CAD~\cite{avetisyan2019scan2cad} & 2019 & 1506 & Synthetic & Indoor & Clutter \& Occlusion \& Real noise & Multi-instance registration \\
        \hline
        WHU-TLS~\cite{dong2020whu-tls} & 2020 & 115 & LiDAR & Outdoor & Point density \& Clutter \& Occlusion & Pairwise coarse \& fine registration\\
        \hline
        3DLoMatch~\cite{huang2021predator} & 2021 & 1781 & Kinect & Indoor & Limited overlap \& Occlusion \& Real noise & Pairwise coarse \& Multi-view registration \\
        \hline
        3DCSR~\cite{huang2021comprehensive} & 2021 & 221 & LiDAR \& Kinect & Indoor & Noise \& Density difference \& Limited overlap & Cross-source registration \\
        \hline
        NSS~\cite{sun2023NNS} & 2023 & 27 & Synthetic & Outdoor & Spatiotemporal changes & Pairwise coarse registration \\
        \hline
        Color3DMatch~\cite{mu2024colorpcr} & 2024 & 1623 & Kinect & Indoor & Occlusion \& Real noise & Color point cloud registration \\
        \hline
        Color3DLoMatch~\cite{mu2024colorpcr} & 2024 & 1781 & Kinect & Indoor & Limited overlap \& Occlusion \& Real noise & Color point cloud registration \\
        \hline
    \end{tabular}}
    \label{Tab:Datasets}
    \end{minipage}
\end{table*}

\subsection{Metrics}
Various evaluation metrics have been proposed to evaluate 3D registration performance. For instance, registration recall (RR), root mean square error (RMSE), mean absolute error (MAE), mean squared error (MSE), mean isotropic error (MIE), rotation error (RE), and translation error (TE) are fundamental metrics for evaluating the accuray performance of most registration tasks. In particular, for the multi-view coarse registration task, maximum correspondence error (MCE) is additionally used to evaluate the maximum displacement of any point on surface from its ground truth position, and empirical cumulative distribution function (ECDF) is used to evaluate the function of the error distribution. 

There are also some metrics used for evaluating some modules in a registration pipeline. For instance, the recall versus precision curve (RPC) is the most frequently used metric for evaluating the performance of local descriptors. Inlier recall (IR), inlier precision (IP), and F1-score (F1) are frequently used metrics for evaluating the performance of correspondence optimization methods. 
\section{Pairwise Coarse Registration}\label{sec:Pairwise Coarse Registration}
\subsection{Geometric Methods}
\begin{figure}
    \centering
    \includegraphics[width=0.75\linewidth]{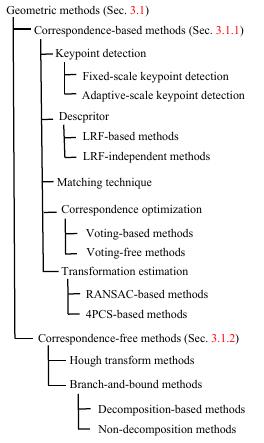}
    \caption{Taxonomy of representative geometric methods for 3D pairwise coarse registration.}
    \label{fig:Taxonomy geometric methods}
\end{figure}

\begin{figure*}[t]
    \centering
    \includegraphics[width=\textwidth]{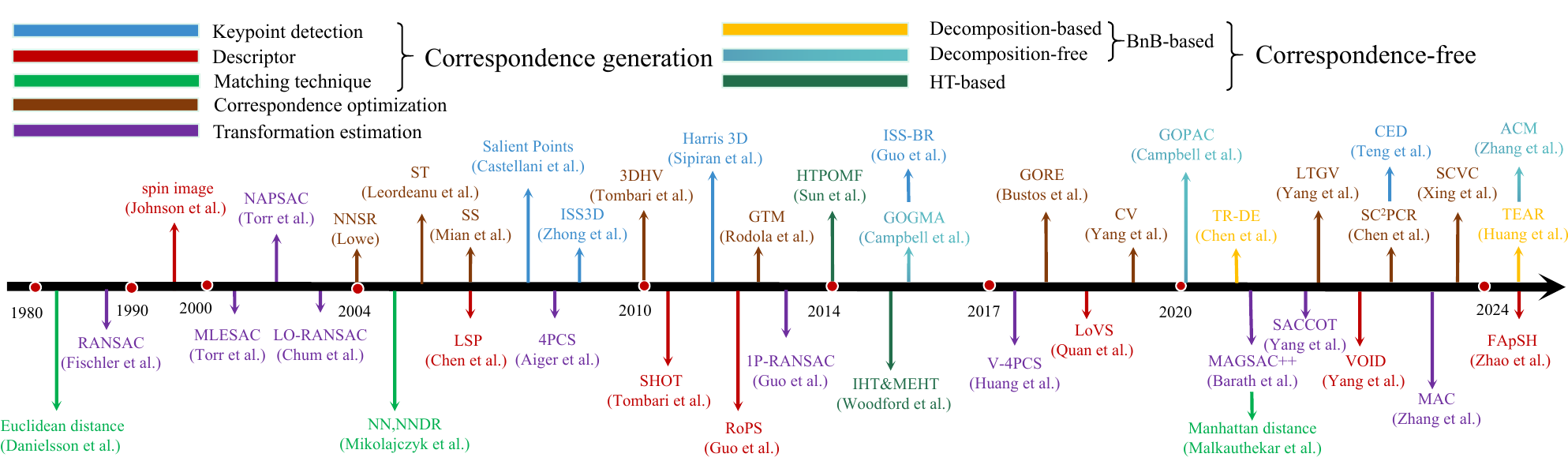}
    \caption{Chronological overview of representative geometric 3D pairwise coarse registration methods.}
    \label{fig:Chronological geometric methods}
\end{figure*}

This section summarizes the pairwise coarse registration methods based on geometric approaches, the taxonomy and chronological overview are shown in Fig.~\ref{fig:Taxonomy geometric methods} and Fig.~\ref{fig:Chronological geometric methods}, respectively. 
\subsubsection{\textbf{Correspondence-based Methods}}
\begin{figure}
    \centering
    \includegraphics[width=1\linewidth]{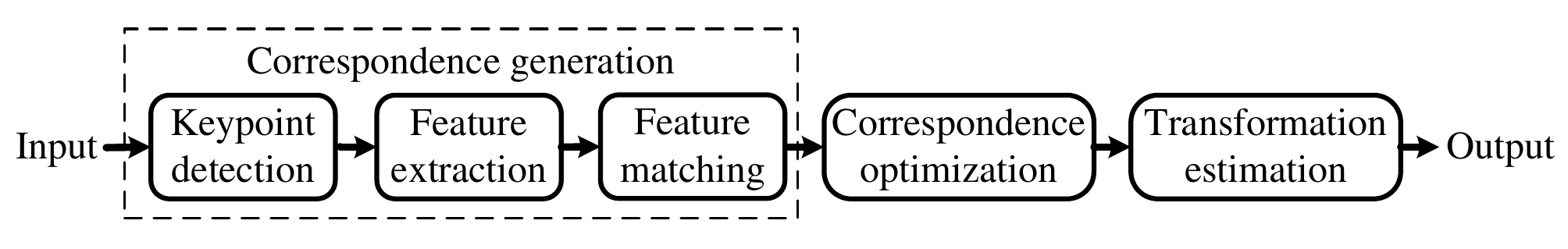}
    \caption{Pipeline of correspondence-based 3D pairwise coarse registration.}
    \label{fig:Correspondence-based}
\end{figure}

In correspondence-based methods, as shown in Fig.~\ref{fig:Correspondence-based}, correspondence generation is crucial for the accuracy and robustness of the registration process.

{\textit{(i)}} \textbf{Keypoint detection}. Its aim is to find a set of sparse yet distinctive points for matching. A comparative evaluation is available in~\cite{tombari2013performance}. A branch of methods detect semantically consistent 3D keypoints~\cite{chen2020unsupervised, jakab2021keypointdeformer,yuan2022unsupervised}, which will be detailed here as they are not specifically designed for registration. We classify existing keypoint detectors for 3D registration as fixed-scale and adaptive-scale.

\textit{1)} \textbf{Fixed-scale keypoint detection methods}. These detectors operate at a fixed scale when identifying keypoints or features across the entire scene, without adjusting the scale based on object size or the distance from the sensor. The main advantage of fixed-scale detectors lies in their faster processing speed and lower computational demands. These methods can be broadly divided into two categories: saliency-based and signature-based. Saliency-based keypoint detection methods~\cite{chen20073d,castellani2008sparse,mian2008keypoint,mian2010repeatability,rusu20113d,teng2023centroid} define a saliency measure to identify regions with distinctive geometric or visual properties, often focusing on local surface variations for efficient keypoint selection. Early works by Chen et al.~\cite{chen20073d} and Castellani et al.~\cite{castellani2008sparse} utilize local surface patches and a 3D saliency measure to detect keypoints in regions with significant surface shape variations. In parallel, Mian et al.~\cite{mian2008keypoint} proposed a method for detecting keypoints with significant shape variations on 3D faces, enabling the extraction of highly descriptive, pose-invariant features. There are also methods that address specific challenges, such as scale invariance~\cite{mian2010repeatability} and computational efficiency~\cite{rusu20113d}. More recently, Teng et al.~\cite{teng2023centroid} introduced the centroid distance (CED) detector, which uniquely identifies keypoints in geometric and color spaces without requiring normal estimation or eigenvalue decomposition. 

Signature-based keypoint detection methods~\cite{sun2009concise,zhong2009intrinsic} rely on specific geometric features to identify keypoints, ensuring robustness and repeatability. For instance, Sun et al.~\cite{sun2009concise} proposed a heat kernel signature (HKS) detector to capture multi-scale neighborhood information, improving stability under shape perturbations. Zhong~\cite{zhong2009intrinsic} developed the intrinsic shape signature (ISS), which provides a view-independent representation of local and semi-local regions for efficient shape matching and pose estimation. There are also several works improving ISS by Guo et al.~\cite{guo2016comprehensive} and Zhang et al.~\cite{zhang2022point}, respectively.
 
Although the above methods are efficient, they generally exhibit limited repeatability performance in complex scenes. 

\textit{2)} \textbf{Adaptive-scale keypoint detection methods}. These detectors dynamically adjust the scale at which keypoints are identified based on local geometry or appearance, ensuring that features are robustly captured across varying object sizes, distances, or levels of detail. An intuitive idea is to extend 2D detectors to 3D. For instance, Sipiran and Bustos~\cite{sipiran2011harris} adapted the widely used 2D Harris operator to 3D, achieving robust keypoint detection by analyzing vertex neighborhoods. Some detectors only adapt to 3D meshes~\cite{zaharescu2009surface,akagunduz20073d,knopp2010hough}. For instance, Zaharescu et al.~\cite{zaharescu2009surface} developed MeshDOG, a scalar field-based method that detects keypoints invariant to rotation, translation, and scale, describing features with local geometric properties. Incorporating additional robustness, Knopp et al.~\cite{knopp2010hough} introduced a method combining local feature extraction with an extended SURF descriptor and a probabilistic Hough transform, significantly improving 3D shape recognition accuracy. Some detectors are designed specifically for 3D point clouds~\cite{novatnack2008scale,unnikrishnan2008multi,steder2011point}, Unnikrishnan and Hebert~\cite{unnikrishnan2008multi} proposed a multi-scale operator approach for direct keypoint detection from raw point clouds, avoiding predefined structures. Steder et al.~\cite{steder2011point} further enhanced keypoint detection by developing the normal aligned radial feature (NARF) method, which integrates object boundary information for better stability and precision.


Although adaptive-scale methods improve robustness against scale variations, their computational complexity sometimes limits real-time efficiency. 

\begin{figure*}[t]
    \centering
    \includegraphics[width=0.8\linewidth]{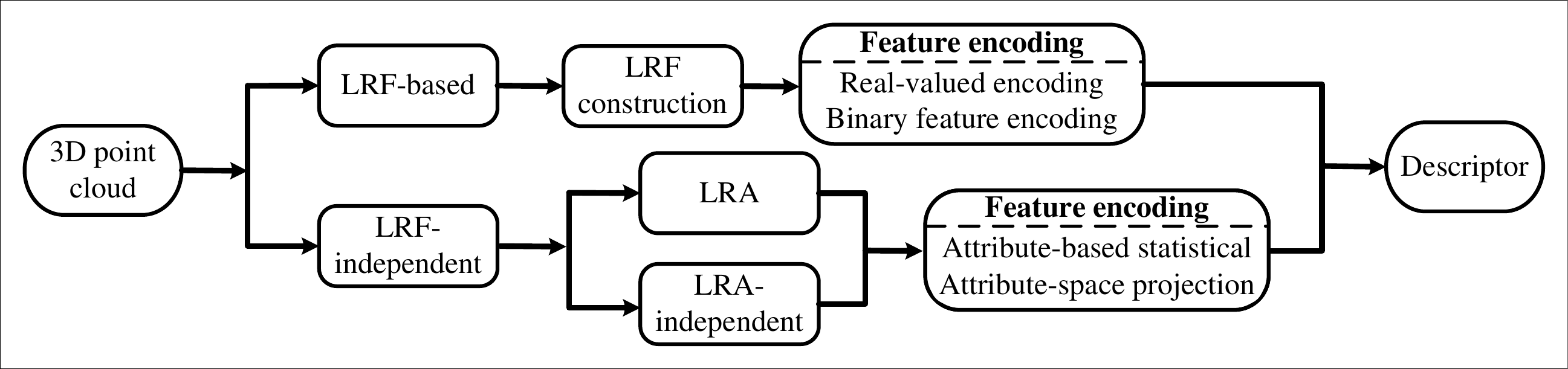}
    \caption{Illustration of 3D local descriptor construction pipeline.}
    \label{fig:Descriptor Framework Overview}
\end{figure*}

{\textit{(ii)}} \textbf{Descriptors}. Once the keypoints are detected, the geometric features of the surrounding local surface can be extracted to generate a local feature descriptor. As shown in Fig.~\ref{fig:Descriptor Framework Overview}, depending on whether a local reference frame (LRF) is established on the local surface of the point cloud, these local feature descriptors can be further classified into two categories: LRF-based and LRF-independent. Table~\ref{Featuredescription} provides an overview of the performance of methods for the description of local surface features.

\textit{1)} \textbf{LRF-based methods}. These descriptors first construct an LRF in the local surface and then encode the spatial and geometric information of the surface based on the LRF. These methods can be further divided into two categories, i.e., real-valued encoding and binary feature encoding. 

Real-valued encoding methods generally generate descriptors with 2D projection attributes, 3D point attributes, or 3D voxel attributes. 
2D-projection-attribute-based methods~\cite{malassiotis2007snapshots,guo2013rotational,yang2017rotational,yang2017toldi,sun2020efficient,liu2023trigonometric,tao2024local,Huang20243d} project the local surface onto 2D planes to efficiently capture geometric distributions by simplifying 3D structures into 2D representations. For instance, Malassiotis and Strintzis~\cite{malassiotis2007snapshots} proposed snapshots, which projects the local surface onto the camera plane, while Guo et al.~\cite{guo2013rotational} developed the rotational projection statistics (RoPS), ensuring a more comprehensive feature encoding with multi-view projection information. Following RoPS, Yang et al.~\cite{yang2017rotational} introduced the rotational contour signatures (RCS), capturing multi-view information through 2D contour projections. By contrast, 3D-point-attribute-based methods~\cite{chen20073d,tombari2010unique,guo2015novel,tombari2011combined,logoglu2016cospair,zhang2021kdd,bibissi2022dual,joshi20243dhonr} utilize spatial and geometric relationships within the 3D neighborhood to describe local surface characteristics. Notable examples include the local surface patch (LSP) representation by Chen et al~.\cite{chen20073d}, which encodes angular relationships into a 2D histogram, and the signature of histograms of orientations (SHOT) by Tombari et al.~\cite{tombari2010unique}, which divides the neighborhood into subspaces for enhanced descriptiveness. 3D voxel attribute methods divide the local region into voxel grids to encode occupancy or density information, effectively representing complex spatial structures. For instance, Tang et al.~\cite{tang2017signature} proposed the signature geometric centroids (SGC) descriptor, which encodes geometry using voxelized centroid-based features. Zhang et al.~\cite{zhang2021kdd} proposed the kernel density-based descriptor (KDD), utilizing kernel density estimation to balance robustness and descriptiveness. 

For binary encoding methods, they focus on simplifying the representation of geometric and spatial information in binary formats, thereby significantly reducing memory usage and accelerating feature matching. We further divide them into descriptor-space binarization and attribute-space binarization methods. For descriptor-space binarization, Prakhya et al.~\cite{prakhya2015b} introduced the binary signature of histograms of orientations (B-SHOT), which uses binary quantization to enhance memory efficiency and matching speed without compromising SHOT's performance. On the other hand, attribute-space binarization methods~\cite{quan2018local,yan2022binary,tao2023distinctive,du2023efficient}, such as the local voxelized structure (LoVS) descriptor proposed by Quan et al.~\cite{quan2018local}, directly compare geometric attribute values within an LRF, improving compactness and efficiency.

LRF-based methods generally exhibit superior descriptiveness performance on clean data, however, their robustness to nuisances such as noise and partial overlap is limited due to LRF's instability.


\textit{2)} \textbf{LRF-independent methods}. LRF-independent descriptors capture local shape information by leveraging geometric properties of the point cloud, such as normal vectors and point densities. These methods can be broadly categorized into two main types: attribute-based statistical methods and attribute-space projection methods. Attribute-based statistical methods~\cite{mitra2003estimating,rusu2008aligning,rusu2009fast,yang2016fast,buch2018local,zhao2020hoppf,zhao2019novel,yang2017multi,zhao2024fapsh} focus on capturing the distribution characteristics of local point clouds through statistical analysis of individual attributes (e.g., distances and angles) or their combinations. These descriptors, such as the point feature histograms (PFH)~\cite{rusu2008aligning}, fast point feature histograms (FPFH)~\cite{rusu2009fast}, use distance relationships to describe local geometry. Some methods encode several attributes simultaneously, such as the local feature statistics histogram (LFSH)~\cite{yang2016fast} and the fully attribute pair statistical histogram (FApSH)~\cite{zhao2024fapsh}. On the other hand, attribute-space projection methods~\cite{johnson1999using,hetzel20013d,ruiz2001new,frome2004recognizing,yang2023void,zhang2024multi,fang2024radiometric} transform the point cloud into an attribute space with structured forms, such as 3D grids or voxel grids. Some methods preject on a 2D attribute space. For instance, spin image~\cite{johnson1999using} and spherical spin image~\cite{ruiz2001new} project local point cloud data onto structured spaces, enabling better handling of occlusions and clutter. Some other methods construct a 3D attribute space for projection. For instance, the voxelization in invariant distance space (VOID) descriptor~\cite{yang2023void} uses voxelization to encode robust features by computing three invariant distance attributes for keypoints and their neighboring points.

LRF-independent methods have recently emerged as a promising and increasingly important trend for 3D local shape description.

{\textit{(iii)}} \textbf{Matching technique}. The main purpose of matching is to accurately identify the corresponding matching pairs of points by comparing the feature descriptors extracted for keypoints. A common approach involves using different distance metrics to perform nearest-neighbor (NN) or nearest-neighbor distance ratio (NNDR) matching. Various distance measures have been used for this purpose, such as the Euclidean distance and the Manhattan distance~\cite{rusu2008aligning}. 

To improve the matching speed, several acceleration techniques have been introduced. Some researchers focus on optimizing the search for nearest neighbors by using data structures such as 2D index tables~\cite{johnson1999using}. Hash tables~\cite{frome2004recognizing,stein1992structural,mian2006three} employ hash functions to quickly map feature descriptors to specific locations, allowing for rapid retrieval. Locality-sensitive trees~\cite{zhong2009intrinsic} group similar data points together to facilitate faster matching, and K-d trees~\cite{guo2013rotational,guo2013trisi,rodola2013scale} are commonly used to accelerate nearest-neighbor searches in high-dimensional spaces. These methods significantly improve the speed and reliability of the matching process.

\begin{table*}[t]
 \centering
  \caption{{Performance summary of representative local feature descriptors.}}
  \vspace{-0.2cm}
  \resizebox*{\textwidth}{!}{
\begin{tabular}{|l|l|c|c|c|}
\hline
Year & Method                                                                                                           & Data Type   & Category & Performance                                            \\ \hline
1999   & Spin image~\cite{johnson1999using}                                                & 
       Mesh \& Point cloud        & LRF-independent   & Rotational invariance and efficient                \\ \hline
2001   & Spherical spin images~\cite{ruiz2001new}                                            
       & Mesh        & LRF-independent   & Outperforms spin image                                 \\ \hline
2004   & 3DSC~\cite{frome2004recognizing}            & Point cloud & LRF-independent   & 
       Outperforms spin image                                 \\ \hline
\multirow{2}{*}{2007}   & LSP~\cite{chen20073d}                                                             & 
       Depth image & LRF-based      & Comparable to spin image, spherical spin image          \\ \cline{2-5}
      & Snapshots~\cite{malassiotis2007snapshots}                                   & 
       Mesh   & LRF-based      & Outperforms spin image                                 \\ \hline
2008   & PFH~\cite{rusu2008aligning}                                                                 & Point cloud & LRF-independent   & Time consuming                                         \\ \hline
2009   & FPFH~\cite{rusu2009fast}                                                           
      & Point cloud & LRF-independent   & More time effcient than PFH                            \\ \hline
2010   & SHOT~\cite{tombari2010unique}                                                    & 
       Mesh        & LRF-based      & Outperforms spin image                                 \\ \hline
2011   & CSHOT~\cite{tombari2011combined}                                                & 
       Mesh        & LRF-based      & Outperforms SHOT                                       \\ \hline
2013  & RoPS~\cite{guo2013rotational}                                                       
      & Mesh        & LRF-based      & Outperforms spin image, LSP, SHOT                        \\ \hline
\multirow{2}{*}{2015}  & B-SHOT~\cite{prakhya2015b}                                                       & Point cloud & LRF-based      & Outperforms SHOT                                       \\ \cline{2-5}
  & TriSI~\cite{guo2015novel}                                                             & Mesh        & LRF-based      & Outperforms spin image, SHOT, RoPS                       \\ \hline
\multirow{2}{*}{2016}  & LFSH~\cite{yang2016fast}                                                          & Point cloud & LRF-independent   & High efficiency and outperforms FPFH, SHOT              \\ \cline{2-5}
  & CoSPAIR~\cite{logoglu2016cospair}                                               & Point cloud & LRF-based      & Outperforms FPFH, SHOT, CSHOT                            \\ \hline
\multirow{3}{*}{2017}  & RCS~\cite{yang2017rotational}                                                     & Point cloud & LRF-based      & Outperforms snapshots, SHOT, B-SHOT, RoPS, RCS             \\ \cline{2-5}
  & TOLDI~\cite{yang2017toldi}                                                          & Point cloud & LRF-based      & Outperforms snapshots, FPFH, SHOT, RoPS                    
  \\ \cline{2-5}
  & MaSH~\cite{yang2017multi}                                                            & Point cloud & LRF-independent      & Outperforms snapshots, FPFH, RoPS, LFSH, RCS               \\ \hline
\multirow{2}{*}{2018}  & LoVS~\cite{quan2018local}                                                            & Point cloud & LRF-based      & Outperforms snapshots, FPFH, SHOT, B-SHOT, RoPS              \\ \cline{2-5}
  & LPPFH~\cite{buch2018local}                                                         & Point cloud & LRF-independent   & Outperforms spin image, FPFH, SHOT, RoPS                  \\ \hline
2019  & SDASS~\cite{zhao2019novel}                                                           & Point cloud & LRF-independent   & Outperforms spin image, FPFH, SHOT, RoPS, LFSH, TriSI, TOLDI \\ \hline
\multirow{2}{*}{2020}  & HoPPF~\cite{zhao2020hoppf}                                                          & Point cloud & LRF-independent   & Outperforms spin image, PFH, FPFH, SHOT, RoPS              \\ \cline{2-5}
  & WHI~\cite{sun2020efficient}                                                          & Point cloud & LRF-based      & Outperforms spin image, SHOT, RoPS, TOLDI, LFSH            \\ \hline
2021  & KDD~\cite{zhang2021kdd}                                                            & Point cloud & LRF-based      & Outperforms FPFH, SHOT, RoPS and short time consuming    \\ \hline
\multirow{4}{*}{2022}  & VOID~\cite{yang2023void}                                                            & Point cloud & LRF-independent   & Outperforms spin image, snapshots, FPFH, SHOT, RoPS, RCS    \\ \cline{2-5}
  & Dual spin-image~\cite{bibissi2022dual}                                           & Point cloud & LRF-based      & Outperforms spin image, snapshots, FPFH, SHOT            \\ \cline{2-5}
 & RSPP~\cite{yan2022binary}                                                            & Point cloud & LRF-based      & Outperforms FPFH, SHOT, RCS, TOLDI                        \\ \hline
\multirow{3}{*}{2023}  & LOVC~\cite{tao2023distinctive}                                                       & Point cloud & LRF-based      & Outperforms spin image, LoVS, RoPS, TOLDI                 \\ \cline{2-5}
 & BWPH~\cite{du2023efficient}                                                            & Point cloud & LRF-based      & Outperforms LoVS, RCS, TOLDI, WHI                         \\ \cline{2-5}

  & TPSH~\cite{liu2023trigonometric}                                                      & Point cloud & LRF-based      & Outperforms spin image, FPFH, SHOT, RoPS, TOLDI            \\ \hline
\multirow{6}{*}{2024}  & RE-LSFH~\cite{fang2024radiometric}                                                 & Point cloud & LRF-independent   & Outperforms FPFH, LFSH                                  \\ \cline{2-5}
  & FApSH~\cite{zhao2024fapsh}                                                          & Point cloud & LRF-independent   & Outperforms spin image, SHOT, RoPS, TOLDI, TriSI, LFSH, MaSH \\ \cline{2-5}
  & LSD~\cite{tao2024local}                                                               & Point cloud & LRF-based      & Outperforms spin image, TOLDI, LoVS                      \\ \cline{2-5}
  & PDSH~\cite{Huang20243d}                                                           & Point cloud & LRF-based      & Outperforms FPFH, SHOT, RoPS, RCS, TOLDI, WHI               \\ \cline{2-5}
  & 3DHoNR~\cite{joshi20243dhonr}                                                       & Point cloud & LRF-based      & High recognition rate, robust to occlusion and noise   \\ \cline{2-5}
  & M-POE~\cite{zhang2024multi}                                                        & Point cloud & LRF-independent   & Outperforms snapshots, FPFH, SHOT, RoPS, RCS, Dual-SI       \\ \hline

    \end{tabular}
  \label{Featuredescription}}
  \vspace{-0.3cm}
\end{table*}

\begin{table*}[t]
\centering
\caption{Comparative results of registration recall for geometric correspondence-based methods on U3M, 3DMatch, 3DLoMatch, KITTI, and ETH datasets. `-' indicates that the result is unavailable. `Trans. Est.' and `Corr. Opt.' refer to transformation estimation and correspondence optimization, respectively.}
\resizebox{\linewidth}{!}{
\begin{tabular}{|l|l|c|c|cc|cc|cc|c|}
\hline
\multirow{3}{*}{Year} & \multirow{3}{*}{Method}                   & \multirow{3}{*}{Category} & U3M        & \multicolumn{2}{c|}{3DMatch}                       & \multicolumn{2}{c|}{3DLoMatch}                     & \multicolumn{2}{c|}{KITTI}                        & ETH                        \\ \cline{4-11} 
                      &                                           &                           & RMSE$<$5pr & \multicolumn{2}{c|}{RE$<$15\textdegree~ TE$<$0.3m} & \multicolumn{2}{c|}{RE$<$15\textdegree~ TE$<$0.3m} & \multicolumn{2}{c|}{RE$<$5\textdegree~ TE$<$0.6m} & RE$<$1\textdegree~ TE$<$1m \\ \cline{4-11} 
                      &                                           &                           & SHOT       & \multicolumn{1}{c|}{FPFH}          & FCGF          & \multicolumn{1}{c|}{FPFH}          & FCGF          & \multicolumn{1}{c|}{FPFH}          & FCGF         & FPFH                       \\ \hline
1981                  & RANASC~\cite{fischler1981random}          & Trans. Est.               & 20.77      & \multicolumn{1}{c|}{64.20}         & 88.42         & \multicolumn{1}{c|}{-}             & 46.38         & \multicolumn{1}{c|}{74.41}         & 98.02        & 73.33                      \\ \hline
2005                  & ST~\cite{leordeanu2005spectral}           & Corr. Opt.                & -          & \multicolumn{1}{c|}{55.88}         & 86.57         & \multicolumn{1}{c|}{-}             & -             & \multicolumn{1}{c|}{-}             & -            & -                          \\ \hline
2009                  & SAC-IA~\cite{rusu2009fast}                & Trans. Est.               & 22.98      & \multicolumn{1}{c|}{-}             & -             & \multicolumn{1}{c|}{-}             & -             & \multicolumn{1}{c|}{-}             & -            & -                          \\ \hline
2016                  & FGR~\cite{zhou2016fast}                   & Corr. Opt.                & 46.77      & \multicolumn{1}{c|}{40.91}         & 78.93         & \multicolumn{1}{c|}{5.09}          &               & \multicolumn{1}{c|}{5.23}          & 89.54        & 18.76                      \\ \hline
\multirow{2}{*}{2018} & GORE~\cite{bustos2017guaranteed}          & Corr. Opt.                & -          & \multicolumn{1}{c|}{-}             & -             & \multicolumn{1}{c|}{-}             & -             & \multicolumn{1}{c|}{-}             & -            & 96.00                      \\ \cline{2-11} 
                      & GC-RANSAC~\cite{barath2021graph}          & Trans. Est.               & -          & \multicolumn{1}{c|}{67.65}         & 92.05         & \multicolumn{1}{c|}{17.01}         & 48.62         & \multicolumn{1}{c|}{78.38}         & 73.69        & 85.33                      \\ \hline
\multirow{2}{*}{2020} & CG-SAC~\cite{quan2020compatibility}       & Trans. Est.               & -          & \multicolumn{1}{c|}{78.00}         & 87.52         & \multicolumn{1}{c|}{-}             & 52.31         & \multicolumn{1}{c|}{74.23}         & 83.24        & -                          \\ \cline{2-11} 
                      & MAGSAC++~\cite{barath2020magsac++}        & Trans. Est.               & -          & \multicolumn{1}{c|}{-}             & -             & \multicolumn{1}{c|}{-}             & -             & \multicolumn{1}{c|}{-}             & -            & 54.67                      \\ \hline
2021                  & TEASER~\cite{yang2020teaser}              & Corr. Opt.                & -          & \multicolumn{1}{c|}{75.48}         & 85.77         & \multicolumn{1}{c|}{35.15}         & 46.76         & \multicolumn{1}{c|}{91.17}         & 94.96        & -                          \\ \hline
\multirow{5}{*}{2022} & Practical O($N^2$)~\cite{li2021practical} & Corr. Opt.                & -          & \multicolumn{1}{c|}{-}             & -             & \multicolumn{1}{c|}{-}             & -             & \multicolumn{1}{c|}{-}             & -            & 100                        \\ \cline{2-11} 
                      & TriVoc~\cite{sun2022trivoc}               & Corr. Opt.                & -          & \multicolumn{1}{c|}{78.42}         & -             & \multicolumn{1}{c|}{37.51}         & -             & \multicolumn{1}{c|}{-}             & -            & -                          \\ \cline{2-11} 
                      & SC$^2$-PCR~\cite{chen2023sc}              & Corr. Opt.                & 39.60      & \multicolumn{1}{c|}{83.73}         & 93.16         & \multicolumn{1}{c|}{38.57}         & 58.73         & \multicolumn{1}{c|}{99.28}         & 97.84        & -                          \\ \cline{2-11} 
                      & SAC-COT~\cite{yang2021sac}                & Trans. Est.               & 48.19      & \multicolumn{1}{c|}{-}             & -             & \multicolumn{1}{c|}{-}             & -             & \multicolumn{1}{c|}{-}             & -            & -                          \\ \cline{2-11} 
                      & TR-DE~\cite{chen2022deterministic}        & Trans. Est.               & -          & \multicolumn{1}{c|}{-}             & 86.99         & \multicolumn{1}{c|}{-}             & 50.4          & \multicolumn{1}{c|}{99.10}         & 98.20        & -                          \\ \hline
\multirow{3}{*}{2023} & GROR~\cite{yan2022new}                    & Corr. Opt.                & -          & \multicolumn{1}{c|}{80.78}         & 92.67         & \multicolumn{1}{c|}{38.52}         & 54.10         & \multicolumn{1}{c|}{-}             & -            & -                          \\ \cline{2-11} 
                      & MV~\cite{yang2023mutual}                  & Corr. Opt.                & -          & \multicolumn{1}{c|}{82.62}         & 93.47         & \multicolumn{1}{c|}{36.16}         & 59.18         & \multicolumn{1}{c|}{98.92}         & 98.20        & -                          \\ \cline{2-11} 
                      & MAC~\cite{zhang20233d}                    & Trans. Est.               & 59.26      & \multicolumn{1}{c|}{84.10}         & 93.72         & \multicolumn{1}{c|}{40.88}         & 59.85         & \multicolumn{1}{c|}{99.46}         & 97.84        & -                          \\ \hline
\multirow{5}{*}{2024} & SUCOFT~\cite{sun2024sucoft}               & Corr. Opt.                & -          & \multicolumn{1}{c|}{85.52}         & -             & \multicolumn{1}{c|}{43.14}         & -             & \multicolumn{1}{c|}{-}             & -            & -                          \\ \cline{2-11} 
                      & G3Reg~\cite{qiao2024g3reg}                & Corr. Opt.                & -          & \multicolumn{1}{c|}{-}             & -             & \multicolumn{1}{c|}{-}             & -             & \multicolumn{1}{c|}{99.46}         & -            & -                          \\ \cline{2-11} 
                      & SCVC~\cite{xing2024efficient}             & Corr. Opt.                & -          & \multicolumn{1}{c|}{89.44}         & 94.43         & \multicolumn{1}{c|}{-}             & 61.30         & \multicolumn{1}{c|}{99.80}         & 100          & -                          \\ \cline{2-11} 
                      & TEAR~\cite{huang2024scalable}             & Trans. Est.               & -          & \multicolumn{1}{c|}{-}             & -             & \multicolumn{1}{c|}{-}             & -             & \multicolumn{1}{c|}{99.10}         & -            & -                          \\ \cline{2-11} 
                      & HERE~\cite{huang2024efficient}            & Trans. Est.               & -          & \multicolumn{1}{c|}{-}             & 91.56         & \multicolumn{1}{c|}{-}             & -             & \multicolumn{1}{c|}{99.10}         & 98.20        & -                          \\ \hline
\end{tabular}}
\label{tab:geometric_registration}
\end{table*}

{\textit{(iv)}} \textbf{Correspondence optimization}. Due to descriptor limitations and challenges such as clutter and occlusions in point cloud data, initial correspondence sets often contain many incorrect matches (outliers). To address this issue, researchers have developed various optimization techniques, generally categorized as voting-based and voting-free. Table~\ref{tab:geometric_registration} provides an overview of the performance for the correspondence optimization methods.

\textit{1)} \textbf{Voting-based methods}. Inspired by the success of the Hough Transform~\cite{illingworth1988survey} in 2D image processing, early 3D methods extend this approach by transforming 3D correspondences into a Hough space for voting. For instance, Tombari and Stefano~\cite{tombari2010object} aimed to detect tight clusters formed by inliers in the Hough space. To further enhance memory efficiency and precision, Woodford et al.~\cite{woodford2014demisting} introduced intrinsic Hough and minimum-entropy Hough, refining vote filtering to make the method more suitable for 3D applications. Xing et al.~\cite{xing2024efficient} proposed single point correspondence voting and
clustering (SCVC), predicting the transformation with a single correspondence while applying Hough voting to determine the remaining degrees of freedom.

In recent years, methods incorporating geometric constraints have been proposed to optimize correspondences. Buch et al.~\cite{buch2014search} presented the search of inliers (SI), which utilizes low-level geometric invariants for local evaluation and covariant constraints for global voting. Based on this, Sahloul et al.~\cite{sahloul2020accurate} presented a two-stage voting scheme with dense evaluation and ranking of local and global geometric consistency. Wu et al.~\cite{wu2022robust} further refined the approach by first generating a geometric consistency point pair voting set using PPF constraints, followed by selecting compatible correspondence triplets to estimate hypothesis poses, resulting in a more robust final pose voting set. 

More recently, research has shifted towards exploiting the consistency between inliers in the graph space. Yang et al.~\cite{yang2019ranking} proposed consistency voting (CV), which evaluates the consistency of each initial correspondence against a predefined voting set based on rigidity and LRF affinity. Expanding on this idea, Yang et al.~\cite{yang2022correspondence} developed loose-tight geometric voting (LTGV), which balances precision and recall by using complementary loose and tight geometric constraints within a dynamic voting scheme. Sun and Deng~\cite{sun2022trivoc} proposed triple-layered voting with consensus maximization (TriVoC), which decomposes the selection of minimal 3-point sets into three layers, each employing efficient voting and correspondence sorting based on pairwise equal-length constraints. Finally, Yang et al.~\cite{yang2023mutual} proposed mutual voting (MV), which models graph nodes and edges as candidates and voters, refining both to achieve more reliable scoring for correspondence evaluation.

\textit{2)} \textbf{Voting-free methods}. Initially, these methods optimize correspondences with feature similarities~\cite{lowe2004distinctive,mian2006three}. Assuming correct correspondences form strongly connected clusters, methods such as spectral technique (ST)~\cite{leordeanu2005spectral} and geometric consistency (GC)~\cite{chen20073d} identify matches through spectral and spatial clustering. Chen et al.~\cite{chen2023sc} proposed SC$^2$-PCR, which employs second-order spatial compatibility for more distinctive clustering at an early stage. Apart from clustering-based methods, some approaches leverage game theory for inlier selection. Albarelli et al.~\cite{albarelli2010game} introduced a game-theoretic framework to achieve fine surface registration based on global geometric compatibility. Building on this, Rodola et al.~\cite{rodola2013scale} introduced game theory matching (GTM), incorporating scale invariance and enhancing robustness in cluttered scenes. 

Other methods focus on finding global optima in the parameter space using techniques such as branch-and-bound (BnB)~\cite{olsson2008branch}. For instance, Bustos and Chin~\cite{bustos2017guaranteed} proposed guaranteed outlier removal (GORE), which uses geometric operations to reject outliers, with BnB as a subroutine. Aiming for efficiency rather than guaranteed global optimality, Li~\cite{li2021practical} defined correspondence matrix (CM) and augmented correspondence matrix (ACM) for tight bounding. To further accelerate GORE, Li et al.~\cite{li2023qgore} presented quadratic-time GORE (QGORE), which employs a voting-based geometric consistency approach for faster upper-bound estimation while maintaining robustness and optimality. 

Recent voting-free methods also aim to achieve global maximum consensus. Zhou et al.~\cite{zhou2016fast} proposed fast global registration (FGR), using the German-McClure loss and Graduated non-convexity (GNC) for non-convex optimization. In graph-theoretic frameworks, methods such as TEASER~\cite{yang2020teaser} and Segregator~\cite{yin2023segregator} use maximum cliques to prune outliers, with segregator incorporating semantic and geometric information. Lusk et al.~\cite{lusk2021clipper} proposed CLIPPER, which relaxes combinatorial constraints for scalable and optimal solutions. Further extending these concepts, Li et al.~\cite{li2024effective} constructed an undirected graph to select preferred correspondences based on the maximum cliques of reliable edges. Qiao et al.~\cite{qiao2024g3reg} proposed G3Reg, which leverages geometric primitives and a pyramid compatibility graph to solve multiple maximum cliques. Rather than focusing solely on the maximum cliques, graph reliability outlier removal (GROR)~\cite{yan2022new} and supercore maximization and flexible thresholding (SUCOFT)~\cite{sun2024sucoft} employ reliability degrees and K-supercore concepts for robust consensus sets. Additionally, Zhang et al.~\cite{zhang2024fastmac} integrated graph signal processing to accelerate the speed of transformation estimation.

Despite these advancements, achieving optimal trade-offs between speed and accuracy remains challenging, particularly in complex scenes. Future work may focus on leveraging multi-order geometric consistency to improve robustness, accuracy, and efficiency.

\begin{figure}
    \centering
    \includegraphics[width=1\linewidth]{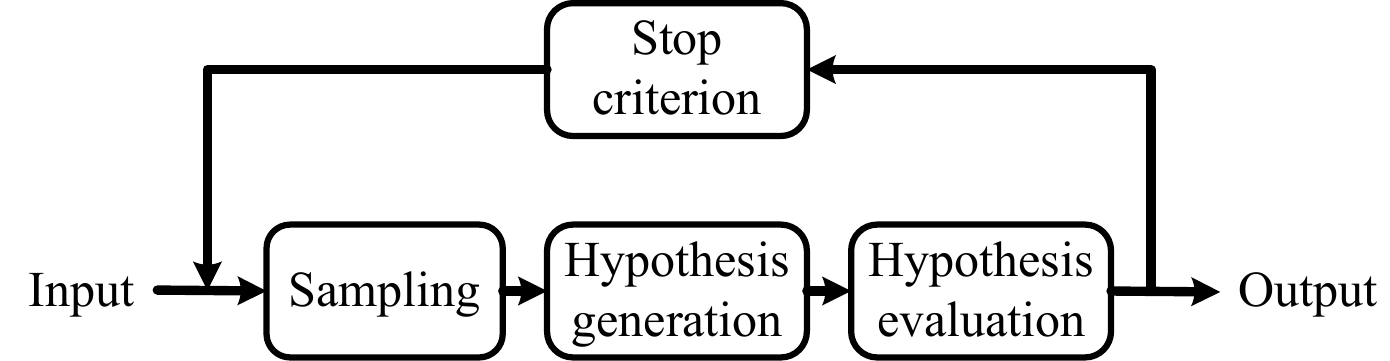}
    \caption{The general pipeline of RANSAC-based methods for 6-DoF pose estimation.}
    \label{fig:ransac}
\end{figure}

{\textit{(v)}} \textbf{Transformation estimation}. After correspondence optimization, the transformation aligning the point clouds is estimated, typically by generating candidate transformations through sampling correspondence subsets, which can be RANSAC- or 4PCS-based. Table~\ref{tab:geometric_registration} provides an overview of the performance for the transformation estimation methods.

\textit{1)} \textbf{RANSAC-based methods}. Random sample consensus~\cite{fischler1981random} (RANSAC) iteratively finds the parameters that fit the input data to a model while discarding outliers, as shown in Fig.~\ref{fig:ransac}. Over the years, many RANSAC variants have emerged, focusing on three main areas of improvement: sampling strategy, evaluation criteria, and local optimization. 

\emph{Sampling Strategy}: Several approaches aim to enhance the sampling process, especially when data is noisy or high-dimensional. Torr et al.~\cite{torr2002napsac} proposed $N$ adjacent points sample consensus (NAPSAC), which prioritizes adjacent points, assuming that inliers are closer together. Barath et al.~\cite{barath2020magsac++} built on this with progressive neighborhood sampling and introduced progressive NAPSAC (P-NAPSAC). Chum and Matas~\cite{chum2005matching} presented progressive sample consensus (PROSAC) that ranks correspondences to enhance the accuracy of match prediction. Following a similar idea, Quan and Yang~\cite{quan2020compatibility} proposed compatibility-guided sample consensus (CG-SAC) to reduce randomness by ranking correspondence pairs based on their compatibility scores. Ni et al.~\cite{ni2009groupsac} introduced GroupSAC to optimize sampling with few inliers by focusing on promising groups. Yang et al.~\cite{yang2021sac} presented sample consensus with compatibility triangles (SAC-COT) that leverages compatibility triangles (COTs) for generating accurate hypotheses. For more flexible constraints, Zhang et al.~\cite{zhang20233d} introduced MAC, which relaxes the previous maximum consensus requirement by using maximal cliques, fully accounting for each local consensus to achieve accurate registration. While a rigid transformation typically requires at least three correspondences~\cite{chen1999ransac}, some methods sample fewer. For instance, Guo et al.~\cite{guo2014integrated} proposed one point RANSAC (1P-RANSAC), which makes a single match with its corresponding LRF sufficient to estimate the pose transformation. Yang et al.~\cite{yang2017multi} proposed 2-point based sample consensus with global constraint (2SAC-GC), which samples two correspondences with associated local reference axes (LRAs). Alternatively, Li et al.~\cite{li2021point} presented one-RANSAC, a method that estimates scaling and translation parameters using a single-point sample without feature information.

\emph{Evaluation Criteria}: RANSAC variants also differ in the way in which they evaluate the quality of a candidate transformation. Torr et al.~\cite{torr2000mlesac} proposed MLESAC, which employs a criterion that maximizes the likelihood rather than just the number of inliers. Chum and Matas~\cite{chum2008optimal} introduced optimal randomized RANSAC (R-RANSAC), which derives the optimality property through a modified sequential probability ratio test (SPRT). Rusu et al.~\cite{rusu2009fast} used the Huber penalty function for evaluation and proposed sample consensus initial alignment (SAC-IA). Addressing limitations in existing metrics, Yang et al.~\cite{yang2021toward} proposed several metrics based on analyzing the contribution of inliers and outliers.

\emph{Local Optimization}: The techniques in this area aim to refine the transformation by locally optimizing it. Chum et al.~\cite{chum2003locally} proposed a locally optimized RANSAC (LO-RANSAC) to reduce the noise brought about by the minimum subset. Barath and Matas~\cite{barath2021graph} introduced graph-cut RANSAC (GC-RANSAC), which applies graph-cut techniques to perform the local optimization step for more precise results. 

Despite efforts to improve RANSAC, these methods often suffer from low efficiency and limited accuracy, especially in extremely low-inlier-ratio cases. Future research could focus on more robust methods to handle high-dimensional data and heavy outliers, while advancements in parallel processing and adaptive sampling could further boost RANSAC’s speed and robustness. 

\textit{2)} \textbf{4PCS-based methods}. The four-point congruent set (4PCS)~\cite{aiger20084} method enhances registration robustness by using point sets with fixed affine ratios, sampling four coplanar points instead of random selections. Variants have been proposed to improve the efficiency: keypoint-based 4PCS (K-4PCS)~\cite{theiler2014keypoint} applies keypoint features to accelerate coplanar set matching, Super4PCS~\cite{mellado2014super} considers angles and uses a 3D grid to limit search areas, and generalized 4PCS (G-4PCS)~\cite{mohamad2014generalized} extends to non-coplanar points. Combining both the advantages of Super4PCS and G-4PCS, Mohamad et al.~\cite{mohamad2015super} proposed super generalized 4PCS, which incorporates adaptive thresholding techniques and dynamic search space constraints. Building on Super4PCS, Huang et al.~\cite{huang2017v4pcs} incorporated volumetric information to further reduce computation time with a volumetric 4PCS (V-4PCS). 

4PCS-based methods are limited by the high computational demands of identifying similar point sets and verifying transformations. Although improved variants offer faster, more accurate, and scalable registration, they still fail for real-time applications. Combining 4PCS with other correspondence optimization techniques could reduce the computational load.

\subsubsection{\textbf{Correspondence-free Methods}}
\begin{figure}
    \centering
    \includegraphics[width=1\linewidth]{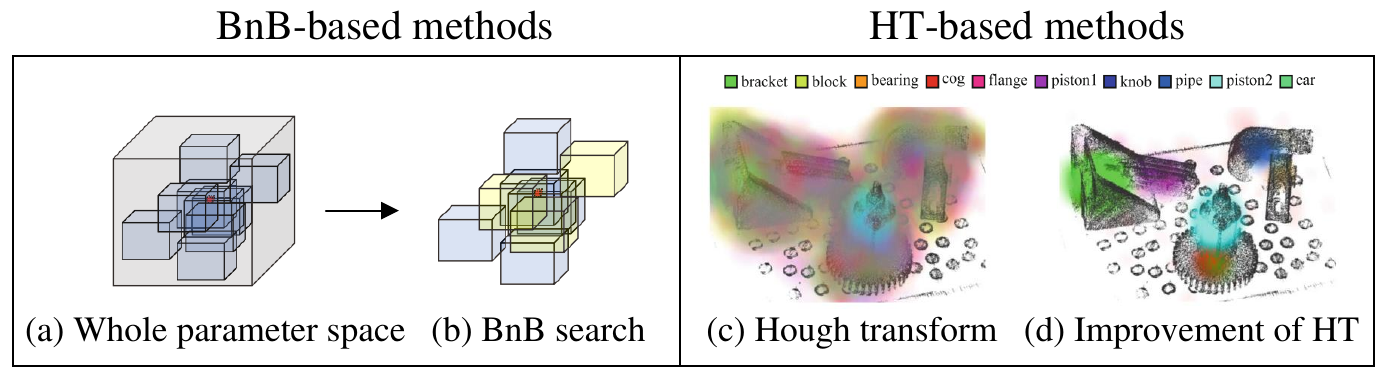}
    \caption{Illustration of typical correspondence-free methods~\cite{2014demisting,2024BnB_heuristics}.}
    \label{fig:Illustration_corr_free}
\end{figure}

These methods perform pairwise coarse registration through optimal parameter search, eliminating the need for correspondence generation or further processing, as shown in Fig.~\ref{fig:Illustration_corr_free}. Furthermore, correspondence-free methods can be broadly categorized into two types, i.e., Hough-transform-based (HT-based) and BnB-based. 

{\textit{(i)}} \textbf{HT-based methods}. The core idea of HT-based methods is to discretize the entire parameter space into a set of bins, subsequently selecting the bin with the highest accumulation of support from the given correspondences as the solution. Hough et al.~\cite{1962HT} first introduced the concept of HT. Then Woodford et al.~\cite{2014demisting} introduced the intrinsic HT that reduces memory and computational requirements, and the minimum-entropy HT to improve precision and robustness, respectively. Sun et al.~\cite{2014HT_mapping} proposed phase-only matched filtering (POMF) for the partially overlapped signal registration, which transforms the input scans into a Hough domain. However, these HT-based methods not only require significant memory and computational requirements, but also may fall into local optima. Consequently, such methods are not widely prevalent.

{\textit{(ii)}} \textbf{BnB-based methods}. Instead of directly searching and voting as the HT-based methods, BnB-based methods recursively partition the parameter space into smaller branches, pruning those suboptimal solutions by assessing the bounds. However, the 6-DoF high-dimensional search space results in exponential time complexity. Depending on the specific acceleration techniques employed, these methods can be broadly classified into decomposition-based and decomposition-free ones.



\textit{1)} \textbf{Decomposition-based methods}. They typically decompose the 6-DOF parameter space into multiple sub-problems, thereby accelerating the BnB process. Liu et al.~\cite{2018rotation_invariant} and Wang et al.~\cite{2021BnB_trans} introduced a rotation invariant feature to decompose the transformation.
 However, the high non-linearity of 3-DoF rotation still limits the efficiency of BnB. Then, Chen et al.~\cite{2022trans_decompo} proposed to decompose the 6-DoF parameter space into (2+1) and (1+2) DoF on rotation axis, an optimal solution is then obtained through a two-stage search strategy. Differently, Huang et al.~\cite{2024tear} employed different strategies, which uses truncated entry-wise absolute residuals and heuristics-guided sampling~\cite{2024BnB_heuristics} to decompose the 6-DoF problem into a set of sub-problems, respectively.
 
 

\textit{2)} \textbf{Decomposition-free methods}. They typically utilize additional constraints to narrow the parameter search space, thereby accelerating the BnB process. Some methods~\cite{2007boxandball,2009GGO,yang2016go,2016gogma} achieve the optimal solution through the combination of BnB and existing technologies. For instance, Li et al.~\cite{2009GGO} reformulated the consensus set maximization problem as a mixed integer programming (MIP) problem, solving it via a tailored BnB method. Yang et al.~\cite{yang2016go} integrated the classic ICP method with the BnB method to solve global optimization problems. Campbell et al.~\cite{2016gogma} presented the GOGMA method employing a BnB approach to solve the problem of 3D rigid Gaussian mixture alignment. In addition, some methods~\cite{2008Euclidean_regis,2014Stereographic_projections,2024ACM} aim to identify more novel and robust constraints. For instance, Olsson et al.~\cite{2008Euclidean_regis} proposed a framework that utilizes various types of correspondences in conjunction with the BnB method to achieve an optimal solution for various pose and registration problems. Bustos et al.~\cite{2014Stereographic_projections} introduced a novel bounding function using stereographic
projections to precompute, and spatially indexed all possible point matches to solve the rotation problem. Recently, Zhang et al.~\cite{2024ACM} proposed a general technique performing a 1-DoF reduction of the space over which BnB is branching to accelerate deterministic consensus maximization. The remaining dimension is solved with an interval stabbing approach. Finally, some methods aim to address specific scenarios through the BnB method. Some BnB methods are also proposed to solve specific application problems. For instance, Cai et al.~\cite{2019terrestrial_LIDAR} presented a fast BnB approach with a polynomial-time subroutine for the 4-DOF scenario of terrestrial LiDAR scan pairs. Campbell et al.~\cite{2018camera_pose} employed the BnB approach to search the 6D space for camera pose and correspondence estimation, and the geometry of SE(3) is used to find upper and lower bounds based on the number of inliers. 

Finally, the performance of correspondence-free methods are shown in Table~\ref{corr-free methods}.

\begin{table*}[t]
  \centering
  \caption{Performance summary of representative correspondence-free 3D pairwise coarse registration methods.}
  \label{corr-free methods}
  \resizebox{\linewidth}{!}{
    \begin{tabular}{|l|l|c|c|c|}
    \hline
    Year & Method & Data Type  & Category & Performance \\
    \hline
    1962 & HT~\cite{1962HT} & Points & HT & Proposes the Hough transform \\
    \hline
    2007 & box-and-ball~\cite{2007boxandball} & Point cloud & BnB & Improvement of BnB using octree\\
    \hline
    2008 & VTPR~\cite{2008Euclidean_regis} & Point cloud & BnB & Solution with cross-type correspondences \\
    \hline
    2009 & BnBMIP~\cite{2009GGO} & Points & BnB & Outperforms RANSAC \\
    \hline
    \multirow{3}{*}{2014} & bnb-M-circ~\cite{2014Stereographic_projections} & Point cloud & BnB & Solve for rotation matrix only \\
    \cline{2-5}
     & IHT\&MEHT~\cite{2014demisting} & CAD models & HT & Improvements in memory requirements and accuracy \\
    \cline{2-5}
     & HTPOMF~\cite{2014HT_mapping} & Point cloud & HT & Outperforms ICP \\
    \hline
    \multirow{2}{*}{2016} & Go-ICP~\cite{yang2016go} & Point cloud & BnB & Suitable for applications with low real-time requirements \\
    \cline{2-5}
     & GOGMA~\cite{2016gogma} & Point cloud & BnB & Outperforms Go-ICP \\
    \hline
    2018 & GoTS~\cite{2018rotation_invariant} & Point cloud & BnB & Outperforms Go-ICP \\
    \hline
    2020 & GOPAC~\cite{2018camera_pose} & CAD models & BnB & Need GPU implementation reducing runtime \\
    \hline
    2021 & GPETD~\cite{2021BnB_trans} & 2D/3D data & BnB & Outperforms GOPAC \\
    \hline
    2022 & TR-DE~\cite{2022trans_decompo} & Point cloud & BnB & Outperforms PointDSC \\
    \hline
    \multirow{3}{*}{2024} & TEAR~\cite{2024tear} & Point cloud & BnB & Suitable for large-scale point pairs \\
    \cline{2-5}
     & HERE~\cite{2024BnB_heuristics} & Point cloud & BnB & Competitive performance with rapid speed \\
    \cline{2-5}
     & ACM~\cite{2024ACM} & Point cloud & BnB & A general strategy to speed up the BnB technique\\
    \hline
    
    \end{tabular}
  \label{resamplingtable}}
\end{table*}

\subsection{Deep-learning-based Methods}
\begin{figure}
    \centering
    \includegraphics[width=0.3\textwidth]{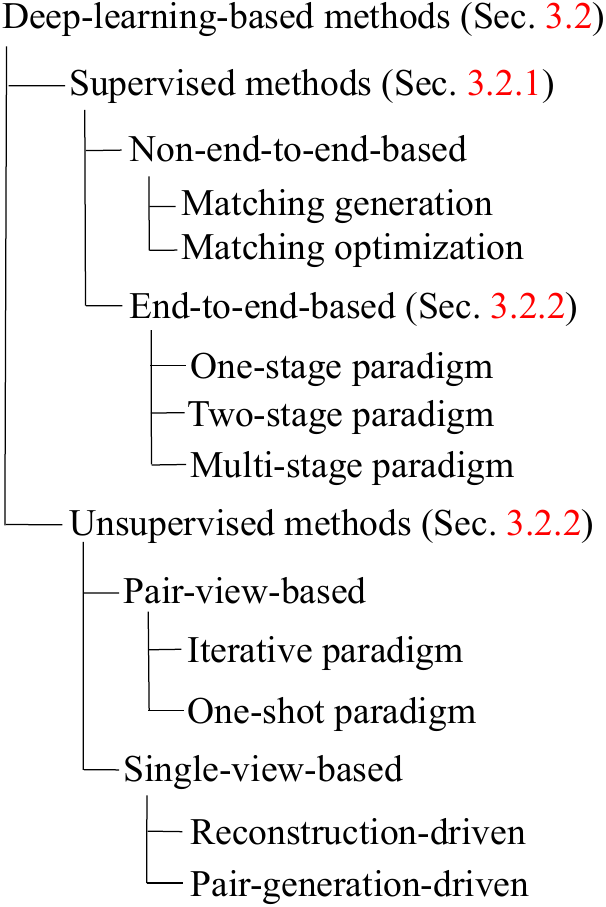}
    \caption{Taxonomy of deep-learning-based 3D pairwise coarse registration methods.}
    \label{fig:DL_reg_cat}
\end{figure}

\begin{figure*}[t]
    \centering
    \includegraphics[width=\linewidth]{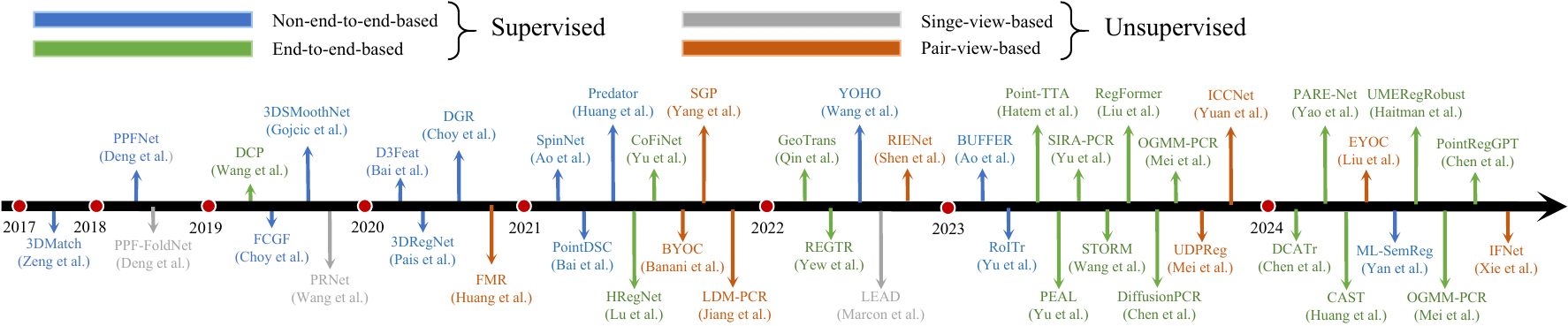}
    \caption{Chronological overview of deep-learning-based 3D pairwise coarse registration methods.}
    \label{fig:DL_reg_time}
\end{figure*}

\begin{table*}[t]
  \caption{Performance comparison of deep-learning-based pairwise coarse registration methods across various datasets. `-' indicates that the result is unavailable. `w' in the `Sup.' column indicates a supervised method, while `w/o' represents unsupervised. `*' signifies omitted digits.}
  \label{tab:DL_performance_comparison}
  \centering
  \resizebox{\textwidth}{!}{
    \begin{tabular}{|c|l|c|c|c|c|c|c|c|c|c|c|c|c|c|c|c|c|c|c|}
    \hline
    \multirow{2}{*}{Year}  & \multirow{2}{*}{Method} & \multirow{2}{*}{Sup.} & \multicolumn{3}{c|}{3DMatch} & \multicolumn{3}{c|}{3DLoMatch} & \multicolumn{3}{c|}{KITTI} & \multicolumn{4}{c|}{ModelNet40-unseen instance} & \multicolumn{4}{c|}{ModelNet40-unseen category} \\
    \cline{4-20}
     & & & FMR (\%) & IR (\%) & RR (\%) & FMR (\%) & IR (\%) & RR (\%) & RTE (cm) & RRE ($^{\circ}$) & RR (\%) & RMSE (R) & RMSE (t) & MAE (R) & MAE (t) & RMSE (R) & RMSE (t) & MAE (R) & MAE (t) \\
    \hline
    2017 & 3DMatch\cite{zeng20173dmatch} & w & 59.6 & - & 67 & - & - & - & - & - & - & - & - & - & - & - & - & - & - \\
    \hline
    \multirow{2}{*}{2018} & PPFNet\cite{deng2018ppfnet} & w & 62.3 & - & 71 & - & - & - & - & - & - & - & - & - & - & - & - & - & - \\
         & PPF-FoldNet\cite{deng2018ppf} & w/o & 71.82 & - & - & - & - & - & - & - & - & - & - & - & - & - & - & - & - \\
    \hline
    \multirow{4}{*}{2020} & DCP\cite{wang2019deep} & w & - & - & - & - & - & - & - & - & - & 1.143385 & 0.001786 & 0.770573 & 0.001195 & 3.150191 & 0.005039 & 2.00721 & 0.003703 \\
         & FCGF\cite{choy2019fully} & w & 95.2 & 56.9 & 87.3 & 76.6 & 21.4 & 40.1 & 4.881 & 0.17 & 97.83 & - & - & - & - & - & - & - & - \\
         & 3DSMoothNet\cite{gojcic2019perfect} & w & 94.7 & 37.7 & 80.3 & 63.6 & 11.4 & 33 & - & - & - & - & - & - & - & - & - & - & - \\
         & PRNet\cite{wang2019prnet} & w/o & - & - & - & - & - & - & - & - & - & 3.199257 & 0.016 & 1.454 & 0.0003 & 3.953 & 0.017 & 1.712 & 0.011 \\
    \hline
    \multirow{8}{*}{2021} & SpinNet\cite{ao2021spinnet} & w & 97.6 & 47.5 & 88.6 & 75.3 & 20.5 & 59.8 & 9.88 & 0.47 & 99.1 & - & - & - & - & - & - & - & - \\
         & PointDSC\cite{bai2021pointdsc} & w & - & 86.54 & 93.28 & - & - & - & 8.13 & 0.35 & 98.2 & - & - & - & - & - & - & - & - \\
         & Predator\cite{huang2021predator} & w & 96.6 & 49.9 & 88.3 & 71.7 & 20 & 54.2 & 6.8 & 0.27 & 99.8 & - & - & - & - & - & - & - & - \\
         & HRegNet\cite{lu2021hregnet} & w & - & - & 93.2 & - & - & - & 4.7 & 0.147 & 100 & - & - & - & - & - & - & - & - \\
         & CoFiNet\cite{yu2021cofinet} & w & 98.1 & 49.8 & 89.3 & 83.1 & 24.4 & 67.5 & 8.5 & 0.41 & 99.8 & - & - & - & - & - & - & - & - \\
         & BYOC\cite{el2021bootstrap} & w/o & 78.6 & - & - & - & - & - & - & - & - & - & - & - & - & - & - & - & - \\
         & SGP\cite{yang2021self} & w/o & - & - & 91.4 & - & - & - & - & - & - & - & - & - & - & - & - & - & - \\
         & LDM-PCR\cite{jiang2021planning} & w/o & - & - & - & - & - & - & - & - & - & 3.0178 & 0.0028 & 0.2779 & 0.00036 & - & - & - & - \\
    \hline
    \multirow{5}{*}{2022} & GeoTransformer\cite{qin2022geometric} & w & 97.9 & 71.9 & 92 & 88.3 & 43.5 & 75 & 7.4 & 0.27 & 99.8 & - & - & - & - & - & - & - & - \\
         & REGTR\cite{yew2022regtr} & w & - & - & 92 & - & - & 64.8 & - & - & - & - & - & - & - & - & - & - & - \\
         & YOHO\cite{wang2022you} & w & 98.2 & 64.4 & 90.8 & 79.4 & 25.9 & 65.2 & - & - & - & - & - & - & - & - & - & - & - \\
         & RIENet\cite{shen2022reliable} & w/o & - & - & - & - & - & - & - & - & - & 0.0033 & 0.0000* & - & - & 0.0059 & 0.0000* & - & - \\
         & LEAD\cite{marcon2021unsupervised} & w/o & 95.84 & - & - & - & - & - & - & - & - & - & - & - & - & - & - & - & - \\
    \hline
    \multirow{11}{*}{2023} & BUFFER\cite{ao2023buffer} & w & - & - & 92.9 & - & - & 71.8 & 5.37 & 0.22 & 97.66 & - & - & - & - & - & - & - & - \\
         & RoITr\cite{yu2023rotation} & w & 98 & 82.6 & 91.9 & 89.6 & 54.3 & 74.8 & - & - & - & - & - & - & - & - & - & - & - \\
         & Point-TTA\cite{hatem2023point} & w & - & - & 93.47 & - & - & 57.81 & - & - & - & - & - & - & - & - & - & - & - \\
         & PEAL\cite{yu2023peal} & w & 99 & 72.4 & 94.6 & 91.7 & 45 & 81.7 & - & - & - & - & - & - & - & - & - & - & - \\
         & SIRA-PCR\cite{chen2023sira} & w & 98.2 & 70.8 & 93.6 & 88.8 & 43.3 & 73.5 & - & - & - & - & - & - & - & - & - & - \\
         & RegFormer\cite{liu2023regformer} & w & - & - & - & - & - & - & 8.4 & 0.24 & 99.8 & - & - & - & - & - & - & - & - \\
         & STORM\cite{wang2022storm} & w & - & - & - & - & - & - & 2.27 & 0.7 & 88.6 & - & - & - & - & - & - & - & - \\
         & OGMM-PCR\cite{mei2023overlap} & w & - & - & - & - & - & - & - & - & - & 0.5892 & 0.0079 & - & - & 0.6309 & 0.0055 & - & - \\
         & DiffusionPCR\cite{chen2023diffusionpcr} & w & 98.3 & 75 & 94.4 & 86.3 & 49.7 & 80 & 6.3 & 0.23 & 99.8 & - & - & - & - & - & - & - & - \\
         & UDPReg\cite{mei2023unsupervised} & w/o & - & - & 91.4 & - & - & 64.3 & - & - & - & - & - & - & - & - & - & - & - \\
         & ICCNet\cite{yuan2024inlier} & w/o & - & - & - & - & - & - & - & - & - & 0.0012 & 0.0000* & - & - & 0.0022 & 0.0000* & - & - \\
    \hline
    \multirow{9}{*}{2024} & DCATr\cite{chen2024dynamic} & w & 98.1 & 76.5 & 92.6 & 87.4 & 48.4 & 76.8 & 6.6 & 0.22 & 99.7 & - & - & - & - & - & - & - & - \\
         & PARE-Net\cite{yao2024pare} & w & 98.5 & 76.9 & 95 & 88.3 & 47.5 & 80.5 & 4.9 & 0.23 & 99.8 & - & - & - & - & - & - & - & - \\
         & CAST\cite{huang2024consistency} & w & - & - & 95.2 & - & - & 75.1 & 2.5 & 0.27 & 100 & - & - & - & - & - & - & - & - \\
         & ML-SemReg\cite{yan2025ml} & w & - & - & - & - & - & - & 5.2 & 0.2 & 98.1 & - & - & - & - & - & - & - & - \\
         & UMERegRobust\cite{haitman2024umeregrobust} & w & - & 79.7 & 93.4 & - & - & - & - & - & - & - & - & - & - & - & - & - & - \\
         & PointRegGPT\cite{chen2024pointreggpt} & w & 98.7 & 71.9 & 93.3 & 89.4 & 45.6 & 77.2 & - & - & - & - & - & - & - & - & - & - & - \\
         & EYOC\cite{liu2024extend} & w/o & - & - & - & - & - & - & - & - & 99.5 & - & - & - & - & - & - & - & - \\
         & RKHS-PCR\cite{zhang2024correspondence} & w/o & - & - & - & - & - & - & - & - & - & 0.02 & - & - & - & - & - & - & - \\
         & IFNet\cite{xie2024iterative} & w/o & - & - & - & - & - & - & - & - & - & 0.0016 & 0.0000* & 0.0007 & 0.0000* & 0.0013 & 0.0000* & 0.0006 & 0.0000* \\
    \hline
    \end{tabular}
  }
\end{table*}

This section summarizes deep-learning-based pairwise coarse registration methods. The taxonomy, chronological overview, and performance comparison are shown in Fig.~\ref{fig:DL_reg_cat}, Fig.~\ref{fig:DL_reg_time} and Tab.~\ref{tab:DL_performance_comparison}.

\subsubsection{\textbf{Supervised Methods}} Supervised learning methods for point cloud registration rely on various types of supervisory signals, such as ground-truth correspondences or transformation parameters, to train models effectively~\cite{lyu2024rigid}. Based on the learning paradigm, the related methods can be divided into non-end-to-end-based and end-to-end-based. The former category focuses on specific stages, such as matching generation or matching optimization, while the latter models the entire point cloud registration process within a single deep learning network.

{\textit{(i)}} \textbf{Non-end-to-end-based methods}. These methods typically focus on specific aspects of point cloud registration, such as feature extraction, correspondence searching, and outlier removal. They can be broadly divided into two categories: matching generation and matching optimization.

\textit{1)} \textbf{Matching generation paradigm}. In this paradigm, the feature extraction network plays a crucial role, which is used to construct the local descriptor for matching keypoints. Similarly to traditional approaches, some methods adopt a patch-wise network that takes local patches as input and outputs descriptors. Typically, a patch-wise network requires a manual method to transform the input local patch into a rotation-invariant space before feature extraction. For instance, some approaches~\cite{khoury2017learning}~\cite{deng2018ppfnet} have enhanced feature compression capabilities by processing high-dimensional hand-crafted descriptors into compact, low-dimensional representations. Some other methods~\cite{zeng20173dmatch}~\cite{poiesi2022learning}~\cite{gojcic2019perfect} utilize LRFs to extract features in a canonical pose. More recent proposals~\cite{ao2021spinnet}~\cite{zhao2024ha}~\cite{zhao2023spherenet} employ rotation-equivalent networks to directly extract features from raw point clouds or voxels without relying on pre-processing.

Although patch-wise networks are effective, their fixed receptive fields can lead to the loss of high-level semantic information. To address this, some works~\cite{choy2019fully}~\cite{wang2022you}~\cite{yu2023rotation} shift to the construction of point-wise descriptors from the entire point cloud as input. However, these methods often bypass the keypoint detection step, and instead randomly sample points for feature description. Such randomness may result in descriptors from non-salient regions, introducing noise into later matching steps. To address this problem, other approaches~\cite{bai2020d3feat}~\cite{yew20183dfeat}~\cite{huang2021predator}~\cite{lu2020rskdd}~\cite{ao2023buffer} integrate keypoint detection and descriptor extraction into a unified framework. These methods jointly predict salient scores and descriptors for all points, and then according to the scores of keypoints, using top-scored points to generate feature correspondences.

\textit{2)} \textbf{Matching optimization paradigm}. In this paradigm, researchers always focus on learning an accurate transformation from input noisy feature correspondences. For instance, Pais et al.~\cite{pais20203dregnet} designed a classification backbone to classify inlier correspondences from the initial matching, followed by~\cite{choy2020deep}~\cite{bai2021pointdsc}. Lee et al.~\cite{lee2021deep} and Jiang et al.~\cite{jiang2023robust} introduced deep learning voting methods to improve outlier rejection. Guo et al.~\cite{guo2023accurate} introduced a second-order consistency into the feature space. Yan et al.~\cite{yan2025ml} combined a semantic segmentation network to establish  multi-level semantic consistency of filtered correspondences. For robust pose estimation, Gao et al.~\cite{gao2024deep} designed a lightweight learning-based pose evaluator to enhance the accuracy of pose selection in low-overlap scenarios.

{\textit{(ii)}} \textbf{End-to-end-based methods}. In these methods, each stage of traditional point cloud registration is represented by distinct modules within the network, enabling it to directly output the estimated transformation. This structure better leverages the potential of deep learning technology. These methods can be roughly divided into one-stage, two-stage, and multi-stage paradigms.

\textit{1)} \textbf{One-stage paradigm}. In the early stage of end-to-end point cloud registration, most methods only search the matching relationship between point clouds once or directly estimate the transformation from the input pair. For end-to-end methods that perform a single matching search, improvements have been made in areas such as feature extraction and matching optimization. For example, some approaches introduce handcrafted descriptors to enhance feature discrimination~\cite{slimani2024rocnet++}, while others incorporate feature interactions between point cloud pairs to improve distinctiveness and matching reliability~\cite{fischer2021stickypillars}~\cite{li2022lepard}~\cite{chen2023rethinking}. Some other works~\cite{cao2021pcam}~\cite{wang2022storm}~\cite{wu2022inenet}~\cite{cao2024sharpgconv} directly predict overlap regions or estimated overlap scores for points to enhance the search for correspondences. To improve the reproducibility of matching points, a few works directly predict the coordinates of corresponding points for transformation estimation\cite{lu2019deepvcp}~\cite{lee2021deeppro}. To enhance the inlier selection process, Zhang et al.~\cite{zhang2022end} and Chen et al.~\cite{chen2023deep} optimized the matching matrix to increase the confidence of inlier correspondences. Other one-stage methods estimate transformations directly based on features. For instance, Deng et al.~\cite{deng20193d} utilized FoldingNet~\cite{yang2018foldingnet} to learn rotation-invariant and rotation-aware descriptors, which were used to search for correspondences and estimate the transformation in feature space. Subsequently, Chen et al.~\cite{chen2022detarnet} and Jiang et al.~\cite{jiang2023center} decoupled the transformation to achieve a more reliable transformation estimation.

\begin{figure}
    \centering
    \includegraphics[width=1\linewidth]{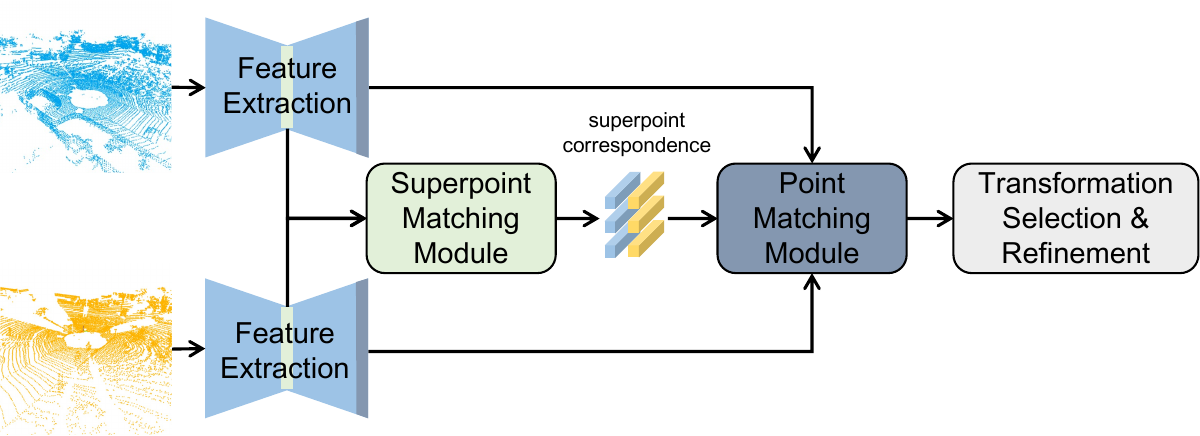}
    \caption{The general framework for two-stage paradigm in supervised 3D pairwise coarse registration methods.}
    \label{fig:DL_registration_sup_two}
\end{figure}

\textit{2)} \textbf{Two-stage paradigm}. The two-stage paradigm often leverages a coarse-to-fine mechanism, whose superiority has been demonstrated in image matching tasks\cite{sun2021loftr}. As shown in Fig.~\ref{fig:DL_registration_sup_two}, these methods typically downsample the input point clouds into superpoints, match them to obtain coarse correspondences, and then propagate these correspondences to individual points based on neighborhood relationships. They finally yield dense point correspondences. 

To our knowledge, CoFiNet~\cite{yu2021cofinet} is the first to introduce this mechanism into point cloud registration, paving the way for subsequent developments. For the coarse-to-fine methods, the superpoint matching accuracy directly determines the final registration performance. Consequently, most methods focus on improving the inlier ratio of coarse correspondences. For instance, GeoTransformer~\cite{qin2022geometric} leverage geometric relationships to enhance the discriminative ability of superpoints, with a set of follow-up works in optimizing superpoint matching~\cite{chen2024fast}~\cite{cao2024dms}~\cite{huang2024consistency}. Other works~\cite{yu2023peal}~\cite{wang2023okr}~\cite{zhao2024lfa}, such as PEAL~\cite{yu2023peal}, incorporated explicit overlap region recognition to more effectively identify inlier correspondences. DCATr~\cite{chen2024dynamic} introduces progressive update mechanisms to improve the spatial consistency of coarse correspondences. PosDiffNet~\cite{she2024posdiffnet}, HRegNet~\cite{lu2021hregnet} and Regformer~\cite{liu2023regformer} introduce  multi-level correspondences to improve the inlier ratio of the coarse matching stage.

In recent years, the two-stage paradigm has been enriched by deep learning innovations from other domains. Chen et al.~\cite{chen2023sira} applied transfer learning to enable generalization from synthetic to real-world scenarios. Hatem et al.~\cite{hatem2023point} proposed a test-time adaptation framework with three self-supervised auxiliary tasks. Yao et al.~\cite{yao2024pare} and Pertigkiozoglou et al.~\cite{pertigkiozoglou2024biequiformer} incorporated rotation-equivariant networks to improve registration performance. Chen et al.~\cite{chen2024pointreggpt} introduced generative models in data augmentation, creating various input pairs to improve training.

\textit{3)} \textbf{Multi-stage paradigm}. Inspired by iterative closest point (ICP), the multi-stage paradigm registers point cloud pairs iteratively, where each iteration constitutes a forward pass of the network. After each pass, the input pairs are aligned using the estimated transformation, which acts as input for the next iteration as well.

Some methods focus on feature-based matching refinement. Wang et al.~\cite{wang2019deep} and Yew et al.~\cite{yew2020rpm} utilized soft matching matrices to estimate transformations, iteratively generating weighted correspondences. Diffusion models have been employed by Jiang et al.~\cite{jiang2024se} and Chen et al.~\cite{chen2023diffusionpcr}, formulating registration as a denoising process to progressively refine transformations.

Some other approaches focus on enhancing feature representations. Wu et al.~\cite{wu2021feature} employed multi-level feature interactions to boost the discriminative power of learned features. Li et al.~\cite{li2023deepsir} proposed a semantic-aware scoring model to leverage both semantic and geometric information, highlighting regions of interest for better registration outcomes. Fu et al.~\cite{fu2021robust} extended this by incorporating deep graph matching and embedding high-order geometric relationships to improve robustness against noise and outliers.

A different line of work bypasses intermediate matching and estimates transformations directly in the feature space. For instance, Aoki et al.~\cite{aoki2019pointnetlk} and Li et al.~\cite{li2021pointnetlk} combined the Lucas \& Kanade algorithm with global feature differences for iterative alignment. Mei et al.~\cite{mei2023overlap} introduced probabilistic registration using Gaussian mixture models (GMMs), while FINet\cite{xu2022finet} designed a dual-branch framework for the direct regression of translation and rotation. 

\subsubsection{\textbf{Unsupervised Methods}} 

The supervised methods mentioned above have shown impressive performance. However, most of their success mainly depends on large amounts of ground-truth transformations between the input pairs, which significantly increases their training costs. Furthermore, obtaining accurate transformation labels can be challenging, due to sensor errors or the reliance on traditional SfM pipelines without convergence guarantees~\cite{schonberger2016structure}. To avoid this, recent efforts have been focused on developing unsupervised registration networks. Based on the input type, unsupervised methods can be roughly divided into two categories: pair-view-based and single-view-based.

{\textit{(i)}} \textbf{Pair-view-based}. These methods always train the network with the pairs of point clouds and design the loss function without ground truth information. We can categorize them into two paradigms for discussion, i.e., iterative and one-shot paradigms.

\begin{figure}
    \centering
    \includegraphics[width=1\linewidth]{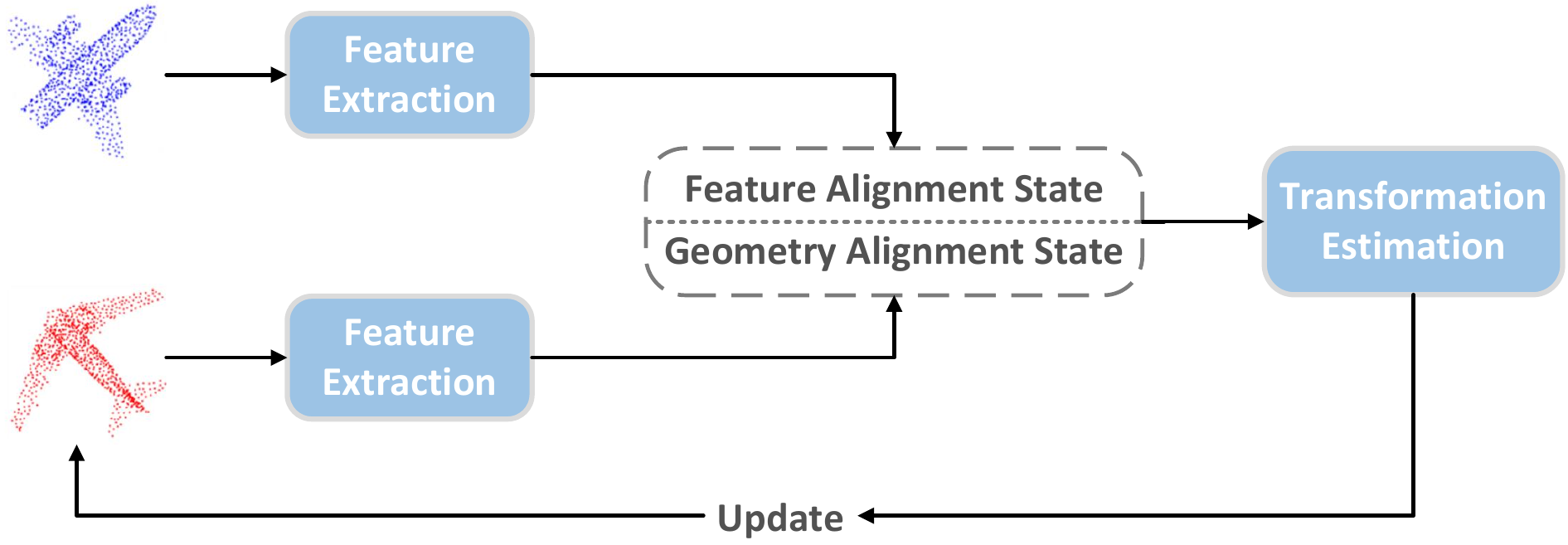}
    \caption{The general framework for iterative paradigm in unsupervised 3D pairwise coarse registration methods.}
    \label{fig:DL_registration_unsup_iterative}
\end{figure}

\textit{1)} \textbf{Iterative paradigm}. these methods are similar to the multi-stage paradigm commonly employed in supervised methods, which perform point cloud registration iteratively during both the training and inference phases. Furthermore, two strategies, i.e., feature alignment and geometric alignment, to align input pairs, as illustrated in Fig.~\ref{fig:DL_registration_unsup_iterative}.

In the feature alignment strategy, researchers ~\cite{huang2020feature}~\cite{liu2023self} ~\cite{jiang2021planning}~\cite{yuan2024learning} typically estimate rigid transformations based on the feature projection error. These methods emphasize that the extracted feature needs to be related to the rigid pose of the input. For example, Huang et al.~\cite{huang2020feature} designed an auto-encoder framework to extract geometry and transformation information, which was later followed by~\cite{liu2023self} and~\cite{jiang2021planning}. Yuan et al.~\cite{yuan2024learning} further maximized multi-hierarchical mutual information in the feature extraction module to obtain discriminative and less redundant representation. Zhang et al.~\cite{zhang2024correspondence} directly leveraged SE(3)-equivariant features for direct feature space registration.

The geometric alignment strategy, on the contrary, requires obtaining point-level correspondences to estimate the transformation using the SVD algorithm. For instance, Jiang et al.~\cite{jiang2021sampling} introduced soft correspondences to generate pseudo-targets and incorporated the geometric alignment error into the loss function. Building on this, Shen et al.~\cite{shen2022reliable} proposed local neighborhood consensus and spatial consistency in pseudo-target generation and loss function definition to improve the ratio of inlier correspondences, followed by~\cite{shi2023overlap}~\cite{yuan2024inlier}~\cite{yu2023amcnet}. Jiang et al.~\cite{jiang2024gtinet} and Zheng et al.~\cite{zheng2024regiformer} combined the feature interaction mechanism between the input pairs for inlier correspondence identification. To improve efficiency, Xie et al.~\cite{xie2024iterative} built on a series of registration blocks. These blocks are cascaded and unfolded recurrently over time to achieve a balance between efficiency and accuracy. Meanwhile, Nie et al.~\cite{nie2024singlereg} performed transformation estimation within a minimal set of correspondences. The iterative paradigm typically yields more precise registration results but demands greater computational resources, which is why most methods focus on object point cloud datasets.

\begin{figure}
    \centering
    \includegraphics[width=1\linewidth]{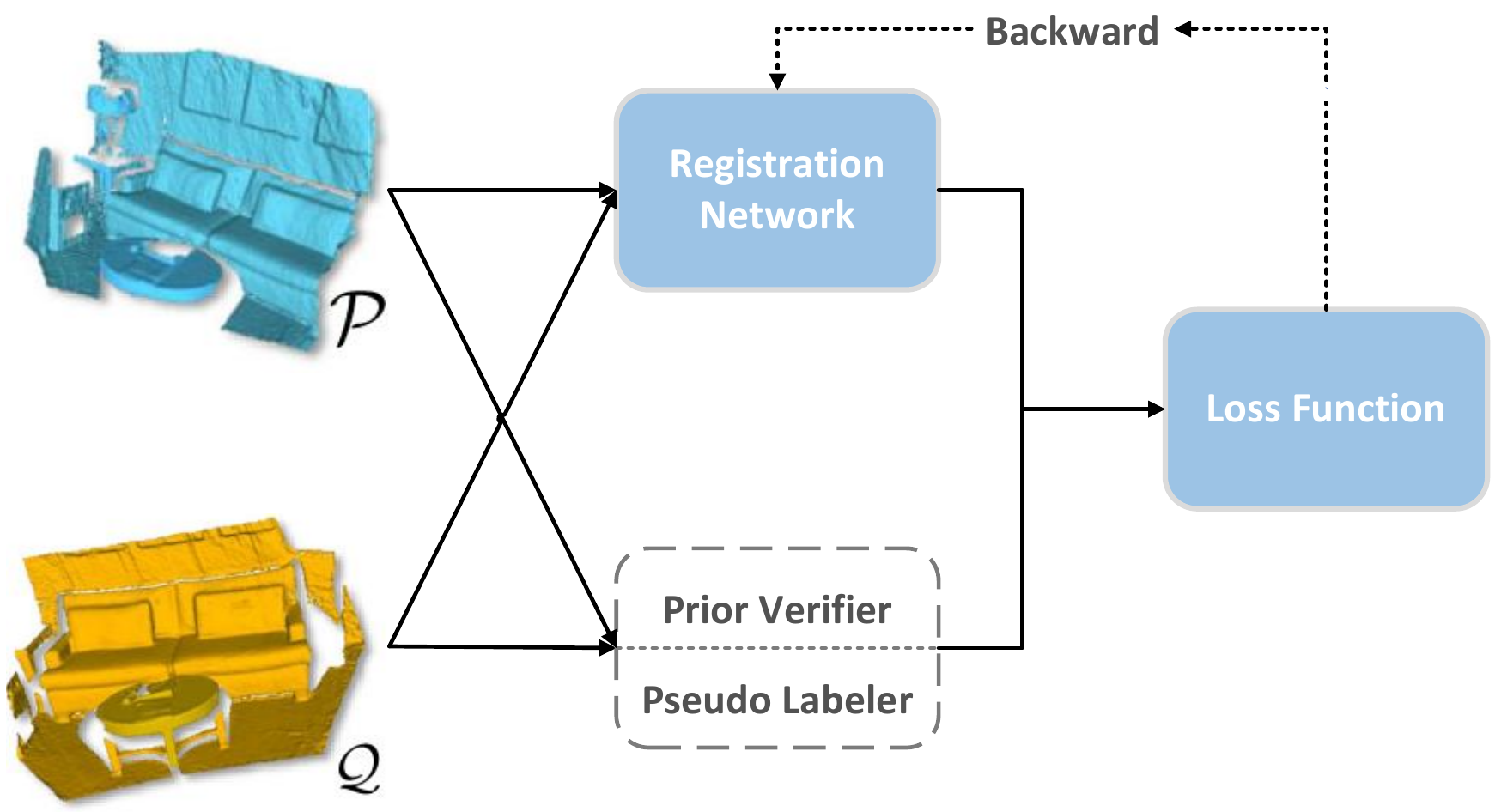}
    \caption{The general framework for one-shot paradigm in unsupervised 3D pairwise coarse registration methods.}
    \label{fig:DL_registration_unsup_onestep}
\end{figure}

\textit{2)} \textbf{One-shot paradigm}. Unlike the iterative paradigm, this paradigm typically registers the input pair only in a one-shot manner. Consequently, these methods always focus on mining prior information or generating pseudo-labels from unlabeled data during the training process, in order to improve registration performance. This is illustrated in Fig.~\ref{fig:DL_registration_unsup_onestep}.

For pseudo-label, Yang et al.~\cite{yang2021self} and Lowens et al.~\cite{lowens2024unsupervised} proposed a teacher-student framework, where the teacher module generates point-wise correspondences as pseudo-label. These labels are then filtered by a verifier to remove low-confidence correspondences before the student module training stage. Liu et al.~\cite{liu2024extend} further extended this framework to distant point cloud registration in a progressive manner.

For prior information, some researchers choose to mine geometric priors. For instance, Kadam et al.~\cite{kadam2022r} introduced LRF to capture transformation-invariant information and constructed matching correspondences at multiple scales based on their PointHop~\cite{zhang2020pointhop} framework. Zhang et al.~\cite{zhang2021representation} trained their network to align the input pair with a canonical pose. Li et al.~\cite{li2022wsdesc} combined the orthogonality and cycle consistency of rigid transformation to guide network training. Huang et al.~\cite{huang2022unsupervised} and Mei et al.~\cite{mei2023unsupervised} utilized GMMs to model the local geometry for evaluating the quality of the registration in the loss function. Other researchers introduce semantic information, such as shape prior and color information. For instance, Hao et al.~\cite{hao20233d} and Li et al.~\cite{li2022unsupervised} obtained the shape prior by part segmentation and shape completion for enhancing object point cloud registration performance. Banani et al.~\cite{el2021bootstrap}~\cite{el2021unsupervisedr} combined RGB-D information to assist in training their indoor point cloud registration network. Compared to the iterative paradigm, the one-shot paradigm is more flexible and better suited for diverse and complex datasets. In recent years, Liu et al.~\cite{liu2024unsupervised} extended the one-shot paradigm to large-scale LiDAR point cloud datasets.

{\textit{(ii)}} \textbf{Single-view-based}: These methods aim to train their networks using only one point cloud as input, which eliminates the need for ``paired data'' and significantly reduces the difficulty of data acquisition. Based on the difference in their training strategies, these methods can be broadly categorized as reconstruction-driven and pair-generation-driven.

\textit{1)} \textbf{Reconstruction-driven}. These methods usually design an auto-encoder framework to reconstruct the input data, thereby learning a compact representation of the input. For example, Deng et al.~\cite{deng2018ppf} developed PPF-FoldNet based on PPFNet~\cite{deng2018ppfnet} and FoldingNet~\cite{yang2018foldingnet}. This approach employs an encoder-decoder framework, where the encoder takes the PPF representation~\cite{birdal2015point} of local patches as input. With the same folding-based auto-encoder framework, Marcon et al.~\cite{marcon2021unsupervised} employed spherical CNNs as the encoder to directly extract rotation-equivalent local descriptors from the original local patches of key points, ensuring the completeness of the input information.

\textit{2)} \textbf{Pair-generation-driven}. These methods train the registration network by generating appropriate registration pairs from single-view inputs. For instance, Sun et al.~\cite{sun2022weakly} generated the registration pair by applying a random rigid transformation to the input point cloud. Wang et al.~\cite{wang2019prnet} simulated partial overlap scenario through random sampling. However, due to the wide variety of noise, partial overlap, and occlusion in real-world scenarios, the pair-generation paradigm struggles to produce a large number of point cloud pairs for registration under diverse conditions. In recent years, a few works~\cite{horache20213d}~\cite{chen2024pointreggpt}~\cite{yao2024pare} incorporate pair generation into their data augmentation pipelines to expand the dataset size for supervised learning.

\subsection{Summary}
We outline the development and characteristics of pairwise coarse registration methods in 3D point cloud registration as follows.

1) {\textbf {Correspondence-based geometric methods.}} More research attention has been shifted to correspondence optimization and transformation estimation in recent years, compared to that of geometric local feature detectors and descriptors. It is still challenging for existing methods to handle low-inlier ratio problems.

2) {\textbf {Correspondence-free geometric methods.}} These methods typically solve the optimal maximum consensus solution through parameter search. However, the 6-DoF high-dimensional search space results in exponential time complexity. Therefore, the BnB-based methods have increasingly popular owing to their globally optimal solution and effectiveness.

3) {\textbf {Learning-based methods.}} Most existing researches toward this line are supervised, achieving great accuracy improvement. However, improving the generalization ability and fostering unsupervised methods deserve further research.

\section{Pairwise Fine Registration}\label{sec:Pairwise Fine Registration}
\begin{figure}
    \centering 
    \includegraphics[width=0.28\textwidth]{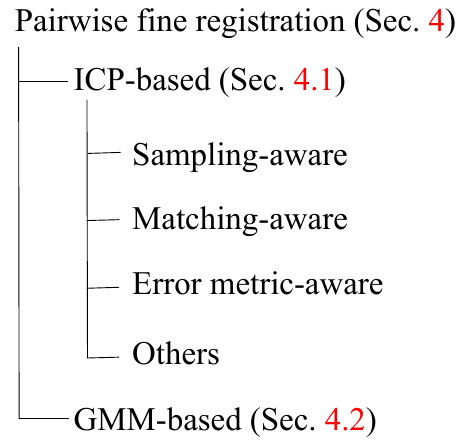}
    \caption{Taxonomy of 3D pairwise fine registration methods.}
    \label{icp1}
    \vspace{-0.45cm}
\end{figure}

\begin{figure*}[t]
    \centering
    \includegraphics[width=\textwidth]{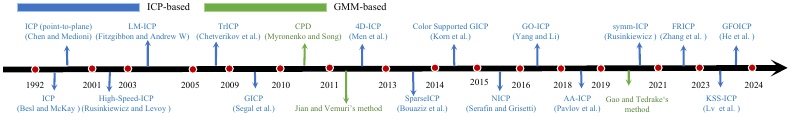}
    \caption{Chronological overview of representative 3D pairwise fine registration methods.}
    \label{icp2}
\end{figure*}

Pairwise fine registration aims to precisely align two point clouds by refining transformations to minimize residual errors. The key methods are based either on ICP or GMMs. This section reviews the main methodologies for pairwise fine registration. The taxonomy, chronological overview, and performance comparison are shown in Fig.\ref{icp1}, Fig.\ref{icp2} and Table.~\ref{icptable}, respectively.

\begin{table*}[t]
  \centering
  \caption{Performance summary of typical 3D pairwise fine registration methods.}
  \vspace{-0.2cm}
  \resizebox*{\textwidth}{!}{
    \begin{tabular}{|c|l|c|c|c|}
    \hline
    Year & Method & Data type & Category & Performance \\
    \hline
    \multirow{2}{*}{1992} & ICP (PTP)~\cite{besl1992method} & Point cloud & ICP-based & Basic method of ICP \\
    \cline{2-5}
    & ICP (PTL)~\cite{chen1992object} & Range images \& Point cloud & Error metric & Basic method of ICP \\
    \hline
    2001 & High-Speed-ICP~\cite{rusinkiewicz2001efficient} & Range images \& Point cloud & Sampling & High efficiency \\
    \hline
    2003 & LM-ICP~\cite{fitzgibbon2003robust} & 2D/3D Point cloud & Error metric & Outperforms ICP \\
    \hline
    2005 & TrICP~\cite{chetverikov2005robust} & 2D/3D Point cloud & Matching & Applicable to cases with low overlap \\
    \hline
    2009 & GICP~\cite{segal2009generalized} & Point cloud & Matching \& Error metric & More robust than ICP \\
    \hline
    2010 & CPD~\cite{myronenko2010point} & Point cloud & GMM-based & Robust to noise, outliers \\
    \hline
    2011 & 4D-ICP~\cite{men2011color} & Point cloud & Matching & High efficiency \\
    \hline
    2013 & SparseICP~\cite{bouaziz2013sparse} & Point cloud & Error metric & Robust to outliers, incomplete data \\
    \hline
    2014 & Color GICP~\cite{korn2014color} & Point cloud & Matching & More robust than GICP \\
    \hline
    2015 & NICP~\cite{serafin2015nicp} & Point cloud & Error metric & Outperforms GICP \\
    \hline
    2016 & GO-ICP~\cite{yang2016go} & Point cloud & Others & Robust to initial estimation \& High efficiency \\
    \hline
    2018 & AA-ICP~\cite{pavlov2018aa} & Point cloud & Error metric & High efficiency \\
    \hline
    2019 & Symmetric-ICP~\cite{rusinkiewicz2019symmetric} & Point cloud & Error metric & Outperforms ICP \\
    \hline
    2021 & FRICP~\cite{zhang2021fast} & Point cloud & Error metric & Outperforms SparseICP \\
    \hline
    \multirow{2}{*}{2023} & KSS-ICP~\cite{lv2023kss} & Point cloud & Others & Robust to similarity transformations \\
    \cline{2-5}
    & GFOICP~\cite{he2023gfoicp} & Point cloud & Sampling \& Error metric & Robust to noise \& Outperforms AA-ICP, ICP \\
    \hline
    \end{tabular}}
  \label{icptable}
\end{table*}

\begin{figure}
    \centering
    \includegraphics[width=1\linewidth]{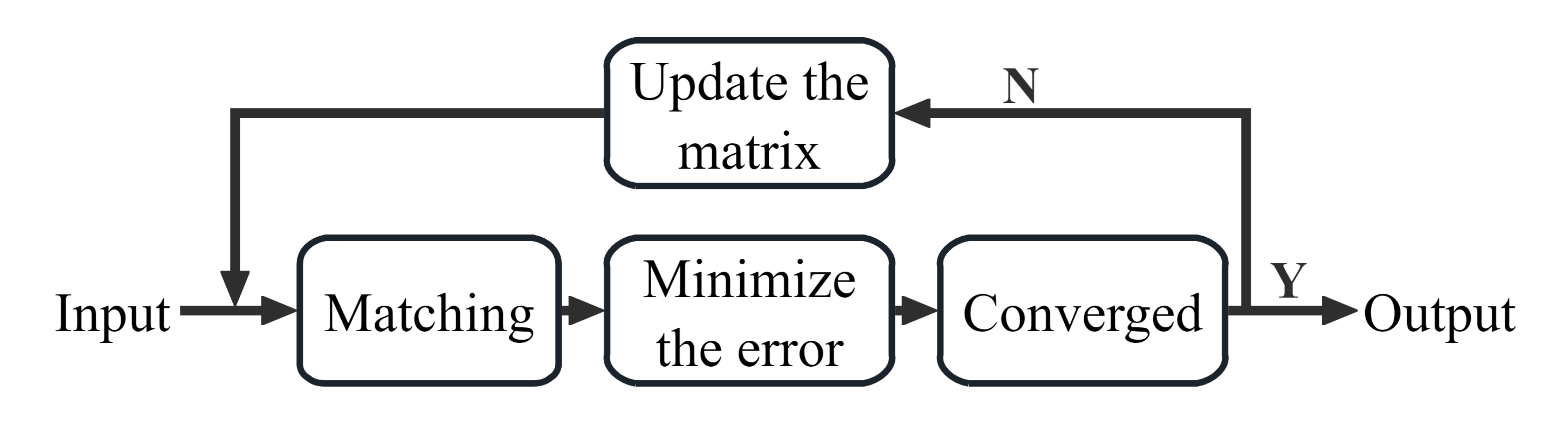}
    \caption{The general pipeline of ICP methods.}
    \label{icp3}
\end{figure}

\subsection{ICP-based Methods}
ICP is first proposed by Besl and McKay~\cite{besl1992method}. It aligns point clouds by iterating between finding matching points and minimizing errors, as illustrated in Fig.~\ref{icp3}. While the standard ICP offers a simple and effective framework, it faces challenges due to its sensitivity to noise, outliers, and the need for good initial alignment. Subsequent improvements attempt to address these issues from different perspectives.

{\textbf{Sampling-aware.}}
In the early stages of ICP development, researchers seek to optimize point selection strategies. Turk and Levoy~\cite{turk1994zippered} incorporated uniform sampling, and Masuda and Yokoya~\cite{masuda1995robust} suggested random sampling. Both sampling methods accelerated the registration process while maintaining registration accuracy. As ICP evolves, sampling-aware methods focus on keypoints selection to reduce erroneous correspondences in the matching stage, such as normal-space sampling~\cite{rusinkiewicz2001efficient} and local-geometric-feature-based sampling in~\cite{he2023gfoicp}.

{\textbf{Matching-aware.}}
Improving the matching module of ICP can effectively enhance robustness against noise and outliers as well as accelerate the registration process. In terms of accelerating the matching process, Benjemaa and Schmitt~\cite{benjemaa1999fast} introduced the multi-z-buffer technique, which optimizes the nearest neighbor search by segmenting 3D space. Men et al.~\cite{men2011color} accelerated matching by incorporating hue-based color information, while color-supported GICP~\cite{korn2014color} introduces color space data to improve matching accuracy and reduce computational load. Some studies focus on enhancing robustness. For example, a statistical method based on distance distribution has been used to improve matching robustness~\cite{zhang1994iterative}. GICP~\cite{segal2009generalized} combines a probabilistic framework with planar structure of scans to enhance correspondences. Additionally, GFOICP~\cite{he2023gfoicp} introduces a distance threshold that gradually shrinks with iterations during matching, and TrICP~\cite{chetverikov2005robust} incorporates the least trimmed squares method. Both methods enhance the robustness of the ICP method by eliminating incorrect correspondences.

{\textbf{Error-metric-aware.}} For ICP methods, the error metric is crucial to determine the alignment quality between two point clouds. A well-designed error metric ensures accurate convergence, reduces misalignment, and handles errors caused by noise or incomplete data. 

One solution is to reduce the impact of incorrect point pairs by assigning different weights, as done in Generalized ICP~\cite{segal2009generalized}, Sparse ICP~\cite{bouaziz2013sparse} and Robust ICP~\cite{zhangg2021fast}. For instance, robust ICP uses the Welsch function to minimize the influence of outliers while maintaining accuracy even in the presence of noise and partial overlap. Another solution involves modifying the error metric to expand the convergence basin or accelerate convergence, as done in methods such as LM-ICP~\cite{fitzgibbon2003robust}, AA-ICP~\cite{pavlov2018aa}, and symmetric ICP~\cite{rusinkiewicz2019symmetric}. LM-ICP and symmetric ICP enhance the registration robustness and accelerate convergence when dealing with smooth and noisy datasets, while AA-ICP leverages historical data to accelerate registration without sacrificing accuracy.

Several methods try to incorporate geometric information into the error metric to improve ICP. For instance, point-to-plane ICP~\cite{chen1992object} utilizes the normal information of point clouds to refine the error metric from point-to-point distance to point-to-plane distance. NICP~\cite{serafin2015nicp} and GFOICP~\cite{he2023gfoicp} incorporate normal vectors and curvature information into the error metric, respectively.

{\textbf{Others.}} The following methods expand on the traditional ICP method, offering broader solutions to overcome specific limitations. Go-ICP~\cite{yang2016go} uses branch-and-bound global optimization within ICP, allowing globally optimal registration throughout the entire SE(3) space, helping to avoid local minima. KSS-ICP~\cite{lv2023kss} leverages Kendall shape space (KSS) to remove translation, rotation, and scaling effects, making it robust for registering point clouds with non-uniform density and noise while preserving shape invariance.

\subsection{GMM-based Methods}
Different to ICPs that rely on deterministic point correspondences, GMM models point sets as probabilistic distributions, framing registration as an optimization of alignment between these distributions. Specifically, GMM methods generally employ the Expectation-Maximization (EM) framework, where the E step estimates probabilistic correspondences, and the M step optimizes transformation parameters by maximizing the expected log-likelihood. This iterative process enables GMM methods to excel in handling noisy or incomplete data, sometimes offering greater flexibility and robustness compared to ICP.

Jian and Vemuri~\cite{jian2005robust}, \cite{jian2010robust} first introduced a method that minimizes the $L2$ distance between two GMMs, achieving robust registration in the presence of noise and outliers. The coherent point drift (CPD) method~\cite{myronenko2010point} performs registration by maximizing the likelihood function, which replaces the $L2$ norm with the KL divergence. Although these methods provide good robustness, they are generally much slower than ICP and almost incapable of scaling to large-scale point clouds, significantly limiting their practicality. To address this issue, Gao and Tedrake~\cite{gao2019filterreg} proposed FilterReg, which incorporates Gaussian filters into the EM framework, reducing computational cost and enabling faster and more robust registration of large-scale point clouds while maintaining accuracy and robustness.

\subsection{Summary}
The following points can be found.

1) {\textbf {ICP remains a standard paradigm.}} ICP methods have formed a standard paradigm for 3D pairwise fine registration. GMM methods are less popular due to their high computational complexity and sensitivity to parameters.

2) {\textbf {Geometric methods are dominating.}} It is interesting that deep learning methods are rare in this area, potentially due to the difficulties in ultra-accurate error prediction.

\section{Multi-view Coarse Registration}\label{sec:Multi-view Coarse Registration}

\begin{figure*}[t]
    \centering
    \includegraphics[width=1\textwidth]{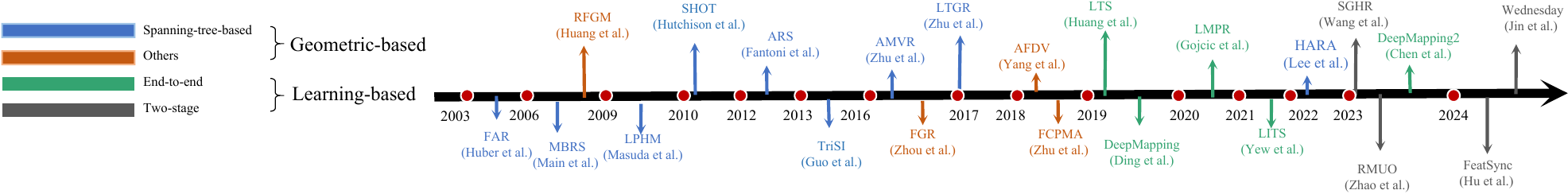}
    \caption{Chronological overview of representative 3D multi-view coarse registration methods.}
    \label{fig:multi coarse}
\end{figure*}

The task of multi-view coarse registration involves aligning multiple point cloud views, captured from different perspectives, into a unified coordinate system using pairwise registration as a foundation.  Compared to pairwise registration, multi-view registration presents additional challenges, such as estimating multiple transformations while addressing cumulative errors and computational overhead. 
The chronological overview,
and performance comparison are shown in Fig.~\ref{fig:multi coarse}
and Table ~\ref{table：multi coarse}, respectively.

\subsection{Geometric Methods}

Geometric methods utilize the intrinsic geometric properties of 3D data to establish correspondence between different views. Based on how these methods construct point cloud topological structures, multi-view point cloud coarse registration geometric methods can be categorized into spanning-tree-based and the others.

{\textbf{Spanning-tree-based}}. 
These methods establish connections through pairwise point cloud registration relationships, unifying the point clouds from different views into the same coordinate system. This is achieved by estimating the relative transformations between nodes to perform the overall registration.

Some methods ~\cite{HUBER2003637,mian2006novel,salti2014shot,masuda2009log} follow a common pipeline, i.e., constructing a spanning tree by exhaustively registering all point cloud pairs.
Specifically, they perform pairwise registration for all point clouds, establish connections for successful registrations to form a connected graph, calculate the spanning tree of the model graph, and finally align the nodes in each connected component using the corresponding transformations. Although these methods can achieve accurate results, their exhaustive nature makes them computationally expensive. To address this issue, several improvements have been proposed. Some approaches involve selecting a set of root nodes based on reliable pairwise registration results and iteratively match them with remaining point clouds ~\cite{guo2013trisi,ZHU20161444,lee2022harahierarchicalapproachrobust,6374979,yang2018aligning}. 
Other methods impose additional constraints on the construction process to reduce the number of pairwise registrations. For example, Zhu et al. ~\cite{zhuLocalGlobalRegistration2017} employed the overlap rate between point clouds as a basis to judge whether a set of point clouds should be registered with each other.
Mian et al.~\cite{mian2006three} and Cheng et al.~\cite{cheng2024incrementalmultiviewpointcloud} performed graph optimization and constructed a multi-level spanning tree hypergraph and a sparse scan graph for acceleration, respectively.

{\textbf{Others}}.
Some methods are based on the perspective of shape enhancement. Guo et al.~\cite{6784345} initialized the seed shape by selecting the view, updated the seed shape sequentially through pairwise registration, and iteratively registered all input views during the shape growth process.
 Zhu el al.~\cite{zhu2018multiviewregistrationunorderedrange}
 employed reliable pairwise registration results to enhance the model shape. 
There are also some optimization-based methods. Huang et al.~\cite{10.1145/1141911.1141925} and Choi et al.~\cite{choi2015robust} both modeled the relative relationships between views as a graph. The former uses a greedy method to iteratively select matching edges with high consistency, merging subgraphs until all views are globally matched. By contrast, the latter uses a global optimization strategy based on line processes to identify and eliminate incorrect pairwise alignments, ultimately achieves a global consistency.
In addition, Zhou et al.~\cite{zhou2016fast} aligned multiple partially overlapping 3D surfaces by directly optimizing a global objective function.

\subsection{Learning-based Methods}
Compared to traditional geometric approaches, learning-based registration methods offer significantly improved computational efficiency while maintaining high registration accuracy. These methods leverage deep neural networks to learn more advanced and robust features, enhancing their effectiveness in diverse scenarios. Based on their implementation strategies, learning-based multi-view point cloud registration methods can be broadly categorized into two types: two-stage and end-to-end.

\begin{table*}[ht]
\centering
\caption{Performance summary of typical 3D multi-view coarse registration methods.}
\vspace{-0.2cm}
\resizebox{\textwidth}{!}{
\begin{tabular}{|c|l|c|c|c|}
\hline
Year & Method                 & Data Type    & Category          & Performance                                                  \\ \hline
2003   & FAR~\cite{HUBER2003637}       & Range image   & Geometric & Sensitive to high symmetry                                   \\ \hline
\multirow{2}{*}{2006}& MBRS~\cite{mian2006three}      & Mesh          & Geometric & Robust to noise                                            
 \\ \cline{2-5}
& RFGM~\cite{10.1145/1141911.1141925}             & Mesh          & Geometric  & Robust to matching features of different sizes              \\ \hline
2009& LPHM~\cite{masuda2009log}             & Range images   & Geometric  & Comparable performance to spin image                       \\ \hline
2010& SHOT~\cite{salti2014shot}             & Mesh          & Geometric  & Outperforms SI, EM, PS                                     \\ \hline
2012& ARS~\cite{6374979}        & Point cloud   & Geometric  & Robust to noise                                             \\ \hline
2013& TriSI~\cite{guo2013trisi}            & Range image    & Geometric  & Outperforms SHOT, Spin image, robust to varying mesh resolutions \\ \hline
\multirow{2}{*}{2016}& AMVR~\cite{ZHU20161444}         & Point cloud   & Geometric  & Outperforms robust to scan orders               
\\ \cline{2-5}
& FGR~\cite{zhou2016fast}              & Point cloud   & Geometric  & Robust to noise and fast in calculation                    \\ \hline
2017& LTGR~\cite{zhuLocalGlobalRegistration2017}   & Mesh          & Geometric  & Outperforms TriSI                                          \\ \hline
\multirow{2}{*}{2018}& AFDV~\cite{yang2018aligning}      & Point cloud   & Geometric  & Robust to data modal changes
\\ \cline{2-5}
& FCPMA~\cite{zhu2018multiviewregistrationunorderedrange}         & Point cloud   & Geometric  & Speed up correspondence propagation                         \\ \hline
\multirow{2}{*}{2019}& LTS~\cite{8954270}              & Point cloud   & Learning   & Outperforms the state-of-the-art transform synchronization techniques on ScanNet and Redwood 
\\ \cline{2-5}
& DeepMapping~\cite{8954379}      & Point cloud   & Learning   & Outperform in scene data \\ \hline
2020& LMPR~\cite{gojcic2020learningmultiview3dpoint}            & Point cloud   & Learning   & Outperforms FGR                                           \\ \hline
2021& LITS~\cite{yew2021learningiterativerobusttransformation}             & Point cloud   & Learning   & Outperforms FGR, LMPR                                     \\ \hline
2022& HARA~\cite{lee2022harahierarchicalapproachrobust}             & Point cloud   & Geometric  & Sensitive to the parameters                                 \\ \hline
\multirow{3}{*}{2023}& SGHR~\cite{wang2023robustmultiviewpointcloud}             & Point cloud   & Learning   & Outperforms LMPR, LITS                                     
\\ \cline{2-5}
& RMUO~\cite{zhao2023registration}            & Point cloud   & Learning   & Comparable performance to SGHR, outperforms FGR            
\\ \cline{2-5}
& DeepMapping2~\cite{10204254}     & Point cloud   & Learning   & Outperforms in scene data            \\ \hline
\multirow{2}{*}{2024}& Wednesday~\cite{jin2024multiwaypointcloudmosaicking}& Point cloud   & Learning   & Outperforms LMPR, LITS, RMPR                              
\\ \cline{2-5}
& FeatSync~\cite{HU2024128088} & Point cloud   & Learning   & Test in scene data                               \\ \hline
\end{tabular}}
\label{table：multi coarse}
\vspace{-0.3cm}
\end{table*}
{\textbf{Two-stage}}.
These methods perform multi-view 3D registration
with two steps or networks comprising two key modules. For the former methods, they generally follow a coarse-to-fine fashion~\cite{wang2023robustmultiviewpointcloud,zhao2023registration,jin2024multiwaypointcloudmosaicking}. For instance, Wang et al.~\cite{wang2023robustmultiviewpointcloud} designed a novel history reweighting function to update the edge weights in the pose graph and employed an iterative reweighted least squares (IRLS) scheme to optimize the residual error.
Jin et al.~\cite{jin2024multiwaypointcloudmosaicking} optimized the pairwise correlation matrices using a diffusion-based procedure, jointly optimizing point cloud rotation and position.
Zhao et al.~\cite{zhao2023registration} constructed a minimum spanning tree by calculating the prediction overlap confidence and leveraged a Lie algebra-based method to optimize the residual error. For the latter methods, they typically involve two sub-networks~\cite{8954379, 10204254}. 
DeepMapping~\cite{8954379}, the first unsupervised method, proposes an L-Net to estimate the pose of the input point cloud, and an M-Net to model the scene structure. It serves the LiDAR mapping problem as a binary classification problem. To cope with large-scale data, DeepMapping2~\cite{10204254}, proposes a place-recognition-based batch organization module to get initial poses, and then introduces a learning-based optimization module for error minimization.

{\textbf{End-to-end}}.
There are still only a few end-to-end methods. As a pioneering work,
Huang et al.~\cite{8954270}
proposed an synchronization method that alternates between transform synchronization and using an iterative neural network.
Gojcic et al.~\cite{gojcic2020learningmultiview3dpoint}
jointly learned initial pairwise alignment and global consistency refinement to improve the robustness of multi-view registration.
Yew et al.~\cite{yew2021learningiterativerobusttransformation}
proposed a graph neural network to learn transformation synchronization and iteratively refined the absolute pose throughout the optimization process. To foster information interaction, Hu et al.~\cite{HU2024128088} recently proposed FeatSync that allows information exchange between different stages, achieving better synchronization result.

\subsection{Summary}
The following points can be summarized.

1) \textbf{Geometry-based coarse registration methods}. These methods mainly rely on geometric features and constraints to achieve registration by constructing topological relationships between pairwise point clouds. While effective in certain scenarios, they struggle with large-scale point clouds, especially when the overlap between views is limited.

2) \textbf{Learning-based coarse registration methods}. Deep learning approaches reduce the reliance on hand-crafted features and heuristic rules, enabling more adaptable and data-driven solutions. However, most existing methods heavily rely on pairwise registration, which limits their generalization capabilities across diverse scenarios. In addition, the sensitivity to outliers can impede the identification of correct connections between views. 

\section{Multi-view Fine Registration}\label{sec:Multi-view Fine Registration}

\begin{figure}
    \includegraphics[width=0.5\textwidth]{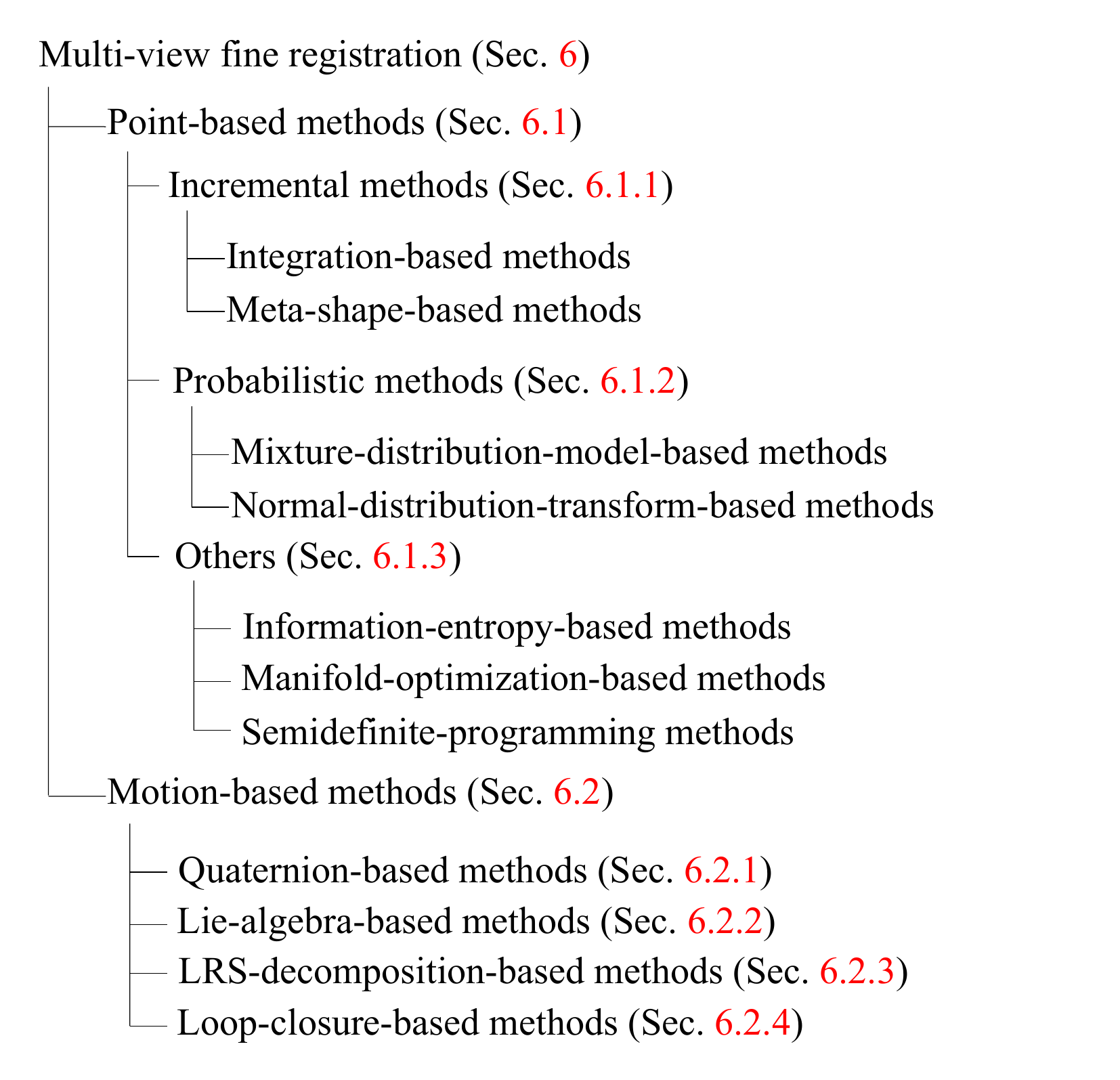}
    \caption{Taxonomy of 3D multi-view fine registration methods.}
    \label{fig:Multi-view Fine Taxonomy}
    \vspace{-0.45cm}
\end{figure}

\begin{figure*}[t]
    \centering
    \includegraphics[width=1\textwidth]{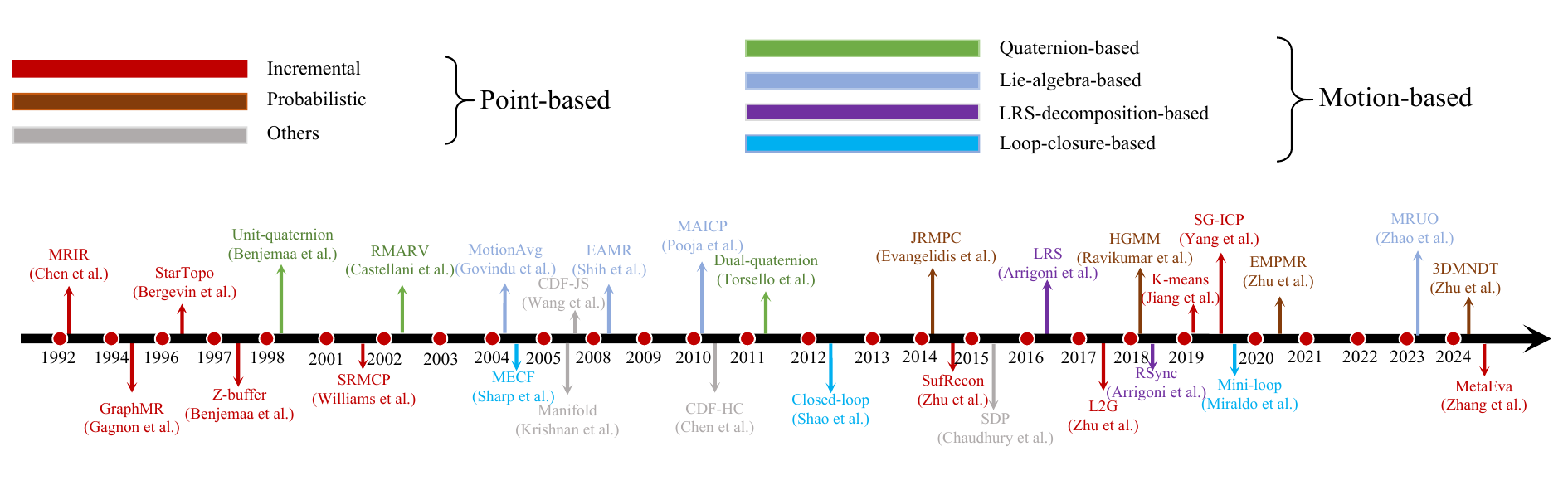}
    \caption{Chronological overview of 3D multi-view fine registration methods.}
    \label{fig:Multi-view_Fine_Chronological}
\end{figure*}

Different from multi-view coarse registration which estimates the transformations of views without initial guesses, multi-view fine registration usually relies on the coarse-aligned transformation input and aims to eliminate the cumulative residual errors. The taxonomy, chronological overview, and performance comparison are shown in Fig.~\ref{fig:Multi-view Fine Taxonomy}, Fig.~\ref{fig:Multi-view_Fine_Chronological} and Table~\ref{tab:Multi_view_Fine_qualitative_methods}, respectively.

\subsection{Point-based Methods}

Point-based methods leverage all available point correspondence information as a constraint to refine the transformation parameters.

\subsubsection{\textbf{Incremental Methods}}
These methods follow an incremental manner that merge single views to a complete point cloud.

{\textit{(i)}} \textbf{Integration-based methods.} These methods sequentially align views to an integrated point cloud. Chen et al.~\cite{chenObjectModellingRegistration1992} proposed a method to merge views sequentially using ICP, but suffers from error accumulation due to the failure of pairwise registration and the lack of global constraints. To address this issue, some methods improve the integration process by establishing a specifically designed graph structure~\cite{gagnonRegistrationMultipleRange1994,pulliMultiviewRegistrationLarge1999}, which can provide a suitable aligned sequence. Bergevin et al.~\cite{bergevinGeneralMultiviewRegistration1996} developed a star-network graph to add global constraints. To further speed up this star-network, some approaches utilize z-buffer segmentation to accelerate point correspondence establishment\cite{benjemaa1997,benjemaa1999fast}. Some methods improve ICP to obtain a higher successful rate during pairwise registration, including color-enhanced~\cite{johnsonRegistrationIntegrationTextured1999} and incorporation of generalized Procrustes analysis~\cite{toldo2010global}. For global constraints, William et al.~\cite{williamsSimultaneousRegistrationMultiple2001} used a constant matrix for direct transformation calculation. Recently, some methods~\cite{fantoniAccurateAutomaticAlignment2012,tang2015hierarchical} improve global alignment c. More recently, Wu et al.~\cite{wuHierarchicalMultiviewRegistration2023} proposed a hierarchical method for scan-to-block integration and model assembly, which shows impressive performance on large-scale datasets.

These methods heavily rely on pairwise fine registration methods. Therefore, they exhibit limited robustness  under low-overlap and noisy conditions.

\begin{figure}
    \includegraphics[width=0.48\textwidth]{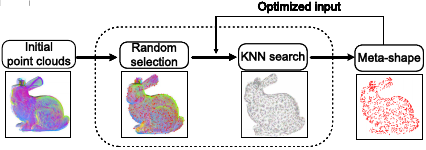}
    \caption{Pipeline of KNN meta-shape method~\cite{zhangMetashapebasedMultiview3D2024}.}
    \label{fig:Meta-shape_pipeline}
\end{figure}

{\textit{(ii)}} \textbf{Meta-shape-based methods.} Similar to integration-based methods, meta-shape-based methods also follow an incremental manner. The main difference between these two categories is that integration-based ones (mentioned as ``meta-view''~\cite{chenObjectModellingRegistration1992,pulliMultiviewRegistrationLarge1999}) merge individual views, which can be seen as ``real-shape'', while meta-shape sample points in each view and merge them into a seed point cloud as a new point cloud. A typical meta-shape-based method pipeline is shown in Fig.~\ref{fig:Meta-shape_pipeline}. Furthermore, meta-shape includes updating operations rather than the simple integration of views.

Zhu et al.~\cite{zhuSurfaceReconstructionEfficient2014} proposed a coarse-to-fine framework that iteratively updates the meta-shape after each view registration. Following this strategy, some methods~\cite{zhuLocalGlobalRegistration2017,yangAligning25DScene2019} also introduce the local-to-global manner to refine the meta-shape. 
There are also some methods~\cite{jiangEfficientRegistrationMultiview2019,guoHierarchicalKmeansClustering2021} update the meta-shape by K-means meta-shapes, employing centroids as transformation estimations.  
Zhang et al.~\cite{zhangMetashapebasedMultiview3D2024} evaluated various meta-shape-based methods, suggesting simple yet effective modifications to improve performance. Recently, Li et al.~\cite{li2024incrementalmultiviewpointcloud} introduced a two-stage candidate retrieval process for meta-shape refinement.

Meta-shape-based methods generally achieve good accuracy performance. However, similar to integration-based methods, their convergence performance heavily depends on their pairwise registration methods used.

\subsubsection{\textbf{Probabilistic Methods}} 
These methods apply probabilistic techniques to address multi-view fine registration, including mixture-distribution-model-based and normal-distribution-transform-based approaches.

{\textit{(i)}} \textbf{Mixture-distribution-model-based methods.} Similar to pairwise fine registration, these methods model point clouds using a mixture distribution, iterating between E-step (expectation) and M-step (maximization) to estimate transformation parameters and update mixture parameters. 

Evangelidis et al.~\cite{evangelidisGenerativeModelJoint2014,evangelidisJointAlignmentMultiple2018} first derived an expectation maximization (EM) manner in multi-view fine registration named JRMPC, which estimates both the GMM parameters and the transformations by mapping individual view onto the ``central'' model. Zhou et al.~\cite{zhouMultiplePointSets2018} adapted GMM to model the point-wise distance. However, these methods are computationally expensive because of the need to estimate numerous parameters and the dependence on initial parameters. Some methods are proposed after JRMPC to further improve the robustness to noise by leveraging other distribution models, such as t-mixture-model (TMM)~\cite{ravikumarGroupwiseSimilarityRegistration2018}, hybrid mixrue model combining Gaussian and Von Mises–Fisher distributions~\cite{minJointRigidRegistration2020}, Laplacian mixture model (LMM)~\cite{zhangRobustMultiviewRegistration2022}. They are demonstrated to be more robust to noise and outliers than GMM. 

To reduce computational overhead, Eckart et al.~\cite{eckartHGMRHierarchicalGaussian2018} introduced a hierarchical Gaussian mixture model, progressively aligning smaller point clouds to optimize the scale of point set correlations. Unlike previous GMM methods such as JRMPC ~\cite{evangelidisJointAlignmentMultiple2018}, which assume all data points are generated from a central GMM, EMPMR~\cite{zhuRegistrationMultiviewPoint2020} assumes that each data point is generated from one corresponding GMM. It only requires to estimate one covariance as well as rigid transformations, which successfully reduces computational cost. In addition, LGS-CPD~\cite{liuLSGCPDCoherentPoint2021} introduces different levels of point-to-plane penalization to add local geometric information to improve GMM performance, which also applies matrix computation on GPU in the E-step for acceleration.

These mixture-based methods offer better robustness as compared with incremental methods, particularly in the presence of outliers. However, the significantly increased computation burden remains an issue.

{\textit{(ii)}} \textbf{Normal-distribution-transform-based methods.} Normal distribution transform (NDT) was introduced by Biber et al.~\cite{biberNormalDistributionsTransform2003} for pairwise 3D registration, which leverages k-means clustering to approximate points in each cluster by normal distributions before registration. Zhu et al.~\cite{zhu3DMNDT3DMultiview2024} extended NDT to multi-view registration, integrating K-means clustering with Lie algebra optimization to enhance registration performance across multiple views.

\begin{table*}[t]
  \centering
  \caption{Performance summary of typical 3D multi-view fine registration methods.}
  \label{tab:Multi_view_Fine_qualitative_methods}
  \vspace{-0.2cm}
  \resizebox{\textwidth}{!}{
    \begin{tabular}{|c|l|c|c|c|}
    \hline
    Year & Method & Data type & Category & Performance \\ \hline
    1992 & MRIR~\cite{chenObjectModellingRegistration1992} & Range image & Integration-based & Tests in object model \\ \hline
    1994 & GraphMR~\cite{gagnonRegistrationMultipleRange1994} & Range image & Integration-based & More balanced graph compared to MRIR; integrates 8 range views of the object into a complete model \\ \hline
    1996 & StarTopo~\cite{bergevinGeneralMultiviewRegistration1996} & Mesh & Integration-based & Integrates up to 20 scans \\ \hline
    1997 & Z-buffer~\cite{benjemaa1997} & Point cloud & Integration-based & Accelerates StarTopo using z-buffer \\ \hline
    1998 & Unit-quaternion~\cite{benjemaa1998solution} & Point cloud & Quaternion-based & Tests in object model \\ \hline
    2001 & SRMCP~\cite{williamsSimultaneousRegistrationMultiple2001} & Point cloud & Integration-based Methods & Outperforms Unit-quaternion \\ \hline
    2002 & RMARV~\cite{castellani2002registration} & Acoustic range view & Quaternion-based & Outperforms MRIR \\ \hline
    \multirow{2}{*}{2004} & MotionAvg~\cite{govindu2004lie} & Data-agnostic & Lie-algebra-based & Outperforms the bundle adjustment method \\ \cline{2-5}
    & MECF~\cite{sharp2004multiview} & Range image & Loop-closure-based & Tests in indoor scene and object model \\ \hline
    2005 & Manifold~\cite{krishnan2005global} & Point cloud & Manifold-optimization-based & Outperforms StarTopo and SRMCP in object data \\ \hline
    2006 & CDF-JS~\cite{wang2006groupwise} & Point cloud & Information-entropy-based & Tests in point cloud data, is immune to noise, and is statistically more robust than the JS \\ \hline
    2008 & EAMR~\cite{shih2008efficient} & Mesh & Lie-algebra-based & Outperforms MotionAvg and MECF in object data \\ \hline
    \multirow{2}{*}{2010} & MAICP~\cite{pooja2010multi} & Point cloud & Lie-algebra-based & Outperforms Z-buffer and MECF in object data \\ \cline{2-5}
    & CDF-HC~\cite{chen2010group} & Point cloud & Information-entropy-based & Outperforms CDF-JS in point cloud data \\ \hline
    2011 & Dual-quaternion~\cite{torsello2011multiview} & Point cloud & Quaternion-based & Outperforms in object data \\ \hline
    2012 & Closed-loop~\cite{shao2012closed} & Point cloud & Loop-closure-based & Tests in generated point cloud data \\ \hline
    \multirow{2}{*}{2014} & JRMPC~\cite{evangelidisGenerativeModelJoint2014} & Point cloud & Mixture-distribution-model-based & Outperforms SRMCP in object data \\ \cline{2-5}
    & SurfRecon~\cite{zhuSurfaceReconstructionEfficient2014} & Point cloud & Meta-shape-based & Outperforms StarTopo and MotionAvg in object data \\ \hline
    2015 & SDP~\cite{chaudhury2015global} & Point cloud & Semidefinite-programming-based & Tests in generated point cloud data \\ \hline
    2016 & LRS~\cite{arrigoni2016global} & Point cloud & LRS-decomposition-based & Outperforms MECF, Dual-quaternion, and MotionAvg in object data \\ \hline
    2017 & L2G~\cite{zhuLocalGlobalRegistration2017} & Point cloud & Meta-shape-based & Outperforms MATrICP in object data \\ \hline
    \multirow{2}{*}{2018} & HGMM~\cite{ravikumarGroupwiseSimilarityRegistration2018} & Point cloud & Mixture-distribution-model-based & Outperforms JRMPC in object data and scene data \\ \cline{2-5}
    & Rsync~\cite{arrigoni2018robust} & Point cloud & LRS-decomposition-based & Outperforms in object data \\ \hline
    \multirow{3}{*}{2019} & K-means~\cite{jiang2019simultaneous} & Point cloud & Meta-shape-based & Outperforms in object data \\ \cline{2-5}
    & SG-ICP~\cite{yangAligning25DScene2019} & Point cloud & Meta-shape-based & Outperforms sequential ICP in object and scene data \\ \cline{2-5}
    & Mini-loop~\cite{miraldo2019minimal} & Point cloud & Loop-closure-based & Tests in scene data \\ \hline
    2020 & EMPMR~\cite{zhuRegistrationMultiviewPoint2020} & Point cloud & Mixture-distribution-model-based & Outperforms MATrICP, JRMPC, LRS, and K-means in object data \\ \hline
    2023 & MRUO~\cite{zhao2023registration} & Point cloud & Lie-algebra-based & Outperforms JRMPC in scene and object data \\ \hline
    \multirow{2}{*}{2024} & MetaEva~\cite{zhangMetashapebasedMultiview3D2024} & Point cloud & Meta-shape-based & Evaluates different variants of meta-shape methods \\ \cline{2-5}
    & 3DMNDT~\cite{zhu3DMNDT3DMultiview2024} & Point cloud & Normal-distribution-transform-based & Outperforms MATrICP, JRMPC, LRS, and K-means in object and scene data \\ \hline
    \end{tabular}
  }
\end{table*}

\subsubsection{\textbf{Others}} In addition to the previously mentioned methods, there are other point-based approaches, such as information-entropy-based, manifold-optimization-based methods, learning-based, and semidefinite-programming-based.

{\textit{(i)}} \textbf{Information-entropy-based methods.} These methods rely on mutual information to measure the similarity between point sets. Methods such as entropy measures based on cumulative distribution functions (CDF) have been proposed to quantify group-wise similarities. These methods offer robust registration even in the presence of noisy or missing correspondences and provide good computational efficiency with closed-form solutions~\cite{wang2006groupwise, chen2010group, sanchez2017group}.

{\textit{(ii)}} \textbf{Manifold-optimization-based methods.} Manifold optimization focuses on transforming the problem of multi-view registration into an optimization problem over a manifold. These approaches ensure that the rotation matrix constraints are maintained during the iterative process. It allows for the alignment of multiple point clouds without static point correspondences, which reduces computational cost and accelerates convergence~\cite{krishnan2005global, krishnan2007optimisation, francescoGlobalRegistrationLarge2014}.


{\textit{(iii)}} \textbf{Semidefinite-programming-based methods.} Semidefinite programming (SDP) is used to relax the least-squares registration problem into a convex optimization, which is easier to solve. These methods leverage SDP formulations to perform global alignment across multiple views, reducing the complexity of the registration process through rank relaxation and iterative methods~\cite{chaudhury2015global, ahmed2017global, ahmed2019least, iglesias2020global}.

\subsection{Motion-based Methods} 

Motion-based (a.k.a transformation synchronization) approaches aim to recover the globally optimal transformation by synchronizing a series of pairwise motions. These methods primarily rely on accurate pairwise transformations. They can be grouped into four main categories: quaternion-based, Lie-algebra-based, LRS-decomposition-based, and loop-closure-based. 

\subsubsection{\textbf{Quaternion-based Methods}} 
Quaternion-based methods leverage the properties of quaternions for iterative optimization of rotation and translation in point cloud registration. Several works employ unit quaternion or dual quaternions for optimizing rotations and transformations. Benjemaa and Schmitt~\cite{benjemaa1998solution} leveraged unit quaternions for optimizing rotations via an iterative process on symmetric $4\times4$ matrices. Fusiello et al.~\cite{fusiello2002model} and Castellani et al.~\cite{castellani2002registration} focused on parameterizing rotation in $SE(3)$ using unit quaternion. 
Torsello et al.~\cite{torsello2011multiview} extended this by using dual quaternions to represent motion and distribute registration errors through a neighbor graph. 
These methods generally aim to improve rotational optimization through the mathematical properties of quaternions, with variations in the form of iterative processes and error distribution strategies.

\subsubsection{\textbf{Lie-algebra-based Methods}} 
These methods leverage Lie group and Lie algebra structures to average transformations and mitigate cumulative registration error. Govindu first introduced motion averaging (MA)~\cite{govindu2004lie} and its RANSAC version~\cite{govindu2006robustness} based on Lie-algebra. Later, Govindu and Pooja~\cite{pooja2010multi,govindu2013averaging} proposed MAICP method that combines MA and ICP and utilizes Lie algebra to compute transformations after obtaining point-wise correspondences in iterations, which is followed by~\cite{zhu2016scaling, guo2018weighted} 

There are also methods that improve global consistency based on graph structure, such as using Dijkstra's algorithm for outlier rejection~\cite{pankaj2017robust}, utilizing maximum connected sub-graph to eliminate unreliable relative motions~\cite{jiang2019simultaneous,jiang20203d}, or leveraging cycle constraints~\cite{shih2008efficient}. In addition, weighting strategies are proposed~\cite{guo2018weighted, bhattacharya2019efficient} to punish unreliable relative transformations.

To mitigate the sensitivity of the F norm to errors in previous methods, Zhu et al.~\cite{zhu2021robust} proposed substituting the F norm for corr-entropy. 
To reduce computational overhead, Bourmaud et al.~\cite{bourmaud2016online} introduced a variational Bayesian approach to deal with large-scale problems. 
Zhao et al.~\cite{zhao2023registration} recently applied Lie algebra as a fine registration block. \par
While Lie-algebra-based methods are efficient, they generally require good initialization. In addition,some approaches only distribute cumulative registration errors equally among views rather than eliminating them.

\begin{figure}
    \includegraphics[width=0.48\textwidth]{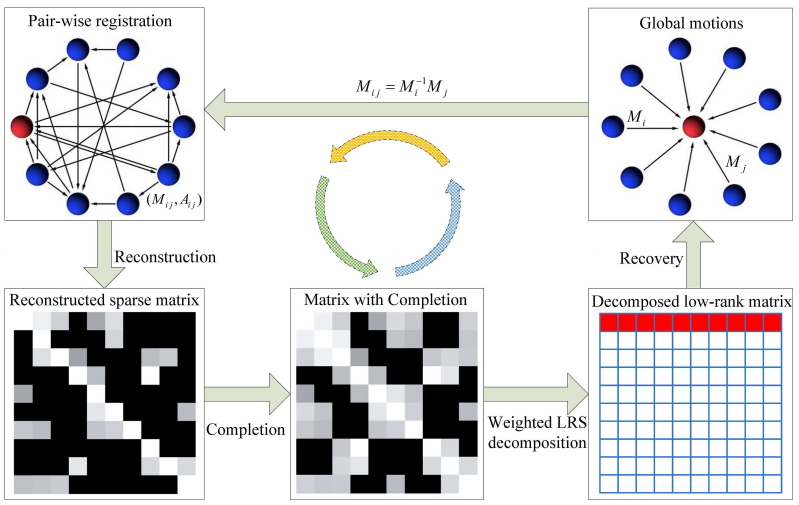}
    \caption{An example of LRS-decomposition-based methods~\cite{jin2019multi}.}
    \label{fig:LRS_structure}
    \vspace{-0.45cm}
\end{figure}

\subsubsection{\textbf{LRS-decomposition-based Methods}} 
Low rank sparsity (LRS) decomposition methods reformulate multi-view registration as an LRS matrix decomposition problem. They optimize transformations by separating them from noise, thereby reducing the impact of noisy data. A typical LRS-decomposition-based method pipeline is shown in Fig.~\ref{fig:LRS_structure}. 
Arrigoni et al.~\cite{arrigoni2016global,arrigoni2018robust} framed registration as an LRS decomposition problem, recovering global motions from a block matrix. However, the reliability of each relative motion differs in reality. LRS assumes that all relative motions have equal reliability, which inevitably leads to accuracy degration. 
To address this issue, several methods propose weighting techniques. Jin et al.~\cite{jin2019multi} assigned the corresponding weights of each scan pair through TrICP for reliable and accurate relative motions. TrICP may run into the local minimum within several iterations, thus Zhang et al.~\cite{zhang2021fast} incorporated angle constraints among point cloud relative motions for weighting.
In addition, Wang et al.~\cite{wang2018multi} introduced a weighted approach from another perspective, which utilizes spatial distribution features of point clouds extracted by spatial rasterization. 
To reduce computational cost, Li et al.~\cite{li2020adaptive} weighted the LRS method utilizing an optimization strategy based on the Lagrange multiplier. It improves registration accuracy and accelerates the process. 

\subsubsection{\textbf{Loop-closure-based Methods}} 
These methods establish the constraints for multi-view point cloud registration through a closed loop and optimize the relative pairwise transformation parameters between views within the closed loop.
Sharp et al.~\cite{sharp2004multiview} defined the problem as an optimization of the graph of neighboring views, and showed that the graph can be decomposed into a set of cycles. Therefore, the optimal transformation parameter for each cycle can be solved in a closed form. Recently, Miraldo et al.~\cite{miraldo2019minimal} also followed this manner that utilizes small loop constraints and fewer point-wise correspondences. To further eliminate pairwise registration error, Liu et al.~\cite{liu2014globally} adapted a parametric bidirectional method to generate reversible transformations in paired overlapping areas thereby eliminating cumulative errors. Some methods~\cite{shao2012closed,shao2014efficient} decouple the rotation matrix and translation vector to distribute cumulative errors.
These methods utilize the loop constraint in connecting graphs to optimize poses, which leads to the same problem as Lie-algebra-based methods that errors are averaged rather than eliminated.

\subsection{Summary}
We outline the development and characteristics of multi-view fine registration methods as follows.

1) {\textbf {The challenge to achieve a performance balance.}} Existing point-based methods deliver good accuracy, while suffer from low time-efficiency. By contrast, motion-based methods are fast but fail to eliminate small errors. Hence, it is still challenging to achieve a balance in terms of accuracy, robustness, and efficiency.

2) {\textbf {The dependence on pairwise ICP.}} The performance of point-based methods usually highly relies on the convergence accuracy of pairwise ICPs. Such dependency may inherit the limitations of ICP methods.

3) {\textbf {Scalability to large-scale unordered data.}} It is still challenging for existing methods to handle a large number of unordered point cloud views.

\section{Other Registration Problems}\label{sec:Other Registration Problems}
Beyond pairwise and multi-view point cloud registration, there are also some other registration problems addressing specific challenges, including cross-scale registration, cross-source registration, color point cloud registration, and multi-instance registration.

\subsection{Cross-scale Point Cloud Registration}
\begin{figure}
    \centering
    \includegraphics[width=0.34\textwidth]{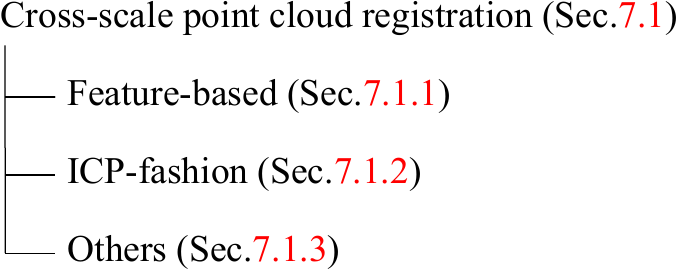}
    \caption{Cross-scale point cloud registration approaches.}
    \label{fig:cross-scale}
    \vspace{-0.45cm}
\end{figure}

\begin{figure*}[t]
    \centering
    \includegraphics[width=1\textwidth]{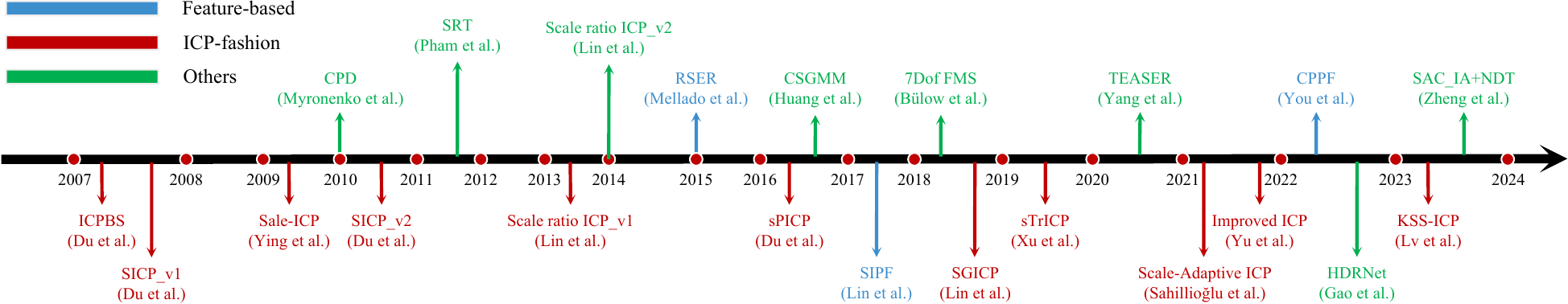}
    \caption{Chronological overview of representative cross-scale 3D point cloud registration methods.}
    \label{fig:cross-scale axis}
\end{figure*}

Cross-scale point cloud registration additionally addresses the scale variation problem. Existing methods are mainly learning-independent, which can be further classified as feature-based, ICP-fashion, and the others. The taxonomy, chronological overview, and performance comparison are shown in Fig.~\ref{fig:cross-scale}, Fig.~\ref{fig:cross-scale axis} and Table~\ref{cross-scale}, respectively.


\begin{table*}[t]
    \centering
    \caption{Performance summary of typical cross-scale 3D point cloud registration methods.}
    \vspace{-0.2cm}
    \resizebox*{\textwidth}{!}{
\begin{tabular}{|c|l|c|c|c|}
\hline
Year & Method               & Data Type             & Category      & Performance                                                            \\ \hline
\multirow{2}{*}{2007} & ICPBS~\cite{du2007ICP}          & 2D shape, point cloud   & ICP-fashion      & Outperforms ICP and ICPS between two m-D point sets                          \\ \cline{2-5} 
                      & SICP\_v1~\cite{du2007Extension}          & 2D shape, point cloud   & ICP-fashion      & General for scaling registration                      \\ \hline
2009   & Scale-ICP~\cite{ying2009Scale}        & Point cloud           & ICP-fashion      & Robust to large-scale stretch and noise                                \\ \hline
\multirow{2}{*}{2010} & CPD~\cite{myronenko2010point}   & Point cloud             & Others        & Suitable for nonrigid point set registration                          \\ \cline{2-5} 
                      & SICP\_v2~\cite{du2010Scaling}          & 2D shape, point cloud   & ICP-fashion      & Independent of shape and features                                      \\ \hline
2011   & SRT~\cite{pham2011New}        & Point cloud           & Others        & High computational cost                                                \\ \hline
2013   & Scale ratio ICP\_v1~\cite{lin2013Scale}         & Point cloud           & ICP-fashion      & High computational cost                                                \\ \hline
2014   & Scale ratio ICP\_v2~\cite{lin2014Scale}         & Point cloud           & Others      & Robust to noise                                                        \\ \hline
2015   & RSER~\cite{mellado2015Relative}     & Multi-modal data      & Feature-based & Outperforms Scale ratio ICP\_v2
     \\ \hline
\multirow{2}{*}{2016} & sPICP~\cite{du2016New}          & Point cloud             & ICP-fashion      & Outperforms ICPBS and CPD                          \\ \cline{2-5} 
                      & CSGMM~\cite{huang2016Coarse}       & Point cloud           & Others        & Effective for large-scale, variable data                               \\ \hline
2017  & SIPF~\cite{lin2018Scale}         & Point cloud           & Feature-based & Robust to scale invariant                                              \\ \hline
\multirow{2}{*}{2018} & 7Dof FMS~\cite{bulow2018Scale}       & Point cloud           & Others        & Robust to noise and partial overlap                          \\ \cline{2-5} 
                      & SGICP~\cite{lin2018SGICP}         & Point cloud           & ICP-fashion      & Enhances GICP with scale variables                                     \\ \hline
2019  & sTrICP~\cite{xu2018Effective}         & 2D shape, range image & ICP-fashion      & Effective for both overlapping and non-overlapping point sets          \\ \hline
2020  & TEASER~\cite{yang2020teaser}        & Point cloud           & Others        & Solves by a simple enumeration                                         \\ \hline
\multirow{2}{*}{2021} & Scale-Adaptive ICP~\cite{sahilliouglu2021Scale} & Point cloud             & ICP-fashion      & Outperforms TEASER                          \\ \cline{2-5} 
                      & Improved ICP~\cite{yu2021Improved}          & Point cloud           & ICP-fashion      & Outperforms ICP, efficient without manual intervention                 \\ \hline
\multirow{2}{*}{2022} & CPPF~\cite{you2022CPPF}          & Point cloud           & Feature-based & Robust to noise                          \\ \cline{2-5} 
                      & HDRNet~\cite{gao2023HDRNet}         & Point cloud           & Others  & Resilient to partial overlaps and noise                                                       \\ \hline
\multirow{2}{*}{2023} & KSS-ICP~\cite{lv2023KSS-ICP}          & Point cloud           & ICP-fashion      & Invariant to similarity transformations                          \\ \cline{2-5} 
                      & SAC\_IA+NDT~\cite{zheng2023Cross}        & Point cloud           & Others        & Outperforms ICP and NDT+ICP                                            \\ \hline
\end{tabular}}
\label{cross-scale}
\end{table*}
\subsubsection{\textbf{Feature-based Methods}}
These methods extract distinctive geometric features from point clouds for the scale variation problem. Some approaches extract local features that are resilient to scale changes and local variations. For instance, Lin et al.~\cite{lin2018Scale} proposed the scale-invariant point feature (SIPF) for keypoint detection, using a voting mechanism to strengthen multi-scale object detection. This approach is further refined through boundary extraction\cite{lin2017Boundary}. To enhance feature selection at varying scales, Lim and Lee~\cite{lim20193D} extended the scale-invariant feature transform (SIFT) to 3D, detecting highly repeatable features and obtaining the support radius regardless of the mesh scale. You et al.~\cite{you2022CPPF} presented the category-level point pair feature (CPPF) for 9D pose estimation, enabling the generalization ability across unseen objects. Some other methods combine local and global features. Mellado et al.~\cite{mellado2015Relative} developed a growing least squares method for scale-invariant matching across noisy, multi-modal data, combining localized accuracy with global structural awareness. 

\subsubsection{\textbf{ICP-fashion Methods}}
These methods minimize distances between point clouds while estimating scale, rotation, and translation parameters. Some approaches~\cite{zinsser2005Point, du2007ICP, du2007Extension, du2010Scaling} integrate scale estimation into the traditional ICP method, extending it to estimate scale alongside rotation and translation. Another category focuses on constraint optimization~\cite{lin2013Scale, ying2009Scale, du2016New, xu2018Effective, sahilliouglu2021Scale}, formulating the registration problem as a high-dimensional optimization task with bounded or adaptive constraints. Additionally, some methods optimize generated correspondences~\cite{lin2018SGICP, wu2019Correntropy, yu2021Improved, chen2022Improved, lv2023KSS-ICP} to enhance the resilience to outliers, noise, and partial overlaps through probabilistic models or advanced similarity measures. 

\subsubsection{\textbf{Others}}
Probabilistic models such as the coherent point drift (CPD) method~\cite{myronenko2010point} leverage GMMs to achieve robust scale alignment in both rigid and nonrigid scenarios. On the other hand, scale characterization techniques, such as those based on principal component analysis (PCA)~\cite{tamaki2010Scale, lin2014Scale}, focus on matching the scale between point clouds, addressing scale variations that hinder accurate alignment. Pham et al.~\cite{pham2011New} further introduced a voting-based technique that uses invariant shape characteristics and distance ratios for scale-independent registration. Transform-based methods, such as the Fourier Mellin SOFT transform~\cite{bulow2018Scale}, address the full degrees of freedom for global alignment. Recent innovations~\cite{lian20233D, zheng2023Cross} improve the robustness of the registration through advanced feature detection and noise filtering. Additionally, some methods focus on outlier rejection~\cite{yang2020teaser, wang2022bounding} to improve registration accuracy and computational efficiency given correspondences. As a pioneering deep-learning-based solution, Gao et al.~\cite{gao2023HDRNet} proposed a high-dimensional regression network HDRNet that effectively handles scale, noise, and partial overlap.

The research toward robust cross-scale registration is at an early stage, and most works assume that the data are fully overlapped. The handling of partial data with scale and rotation variation remains to be solved.

\subsection{Cross-source Point Cloud Registration}
\begin{figure}
    \centering
    \includegraphics[width=0.35\textwidth]{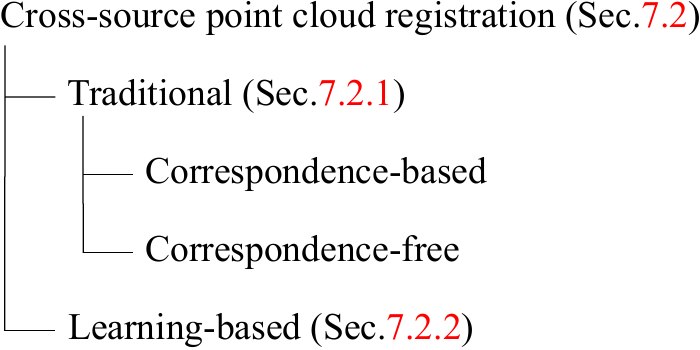}
    \caption{Taxonomy of 3D cross-source point cloud registration approaches.}
    \label{fig:cross-source}
    \vspace{-0.45cm}
\end{figure}

\begin{figure*}[t]
    \centering
    \includegraphics[width=1\textwidth]{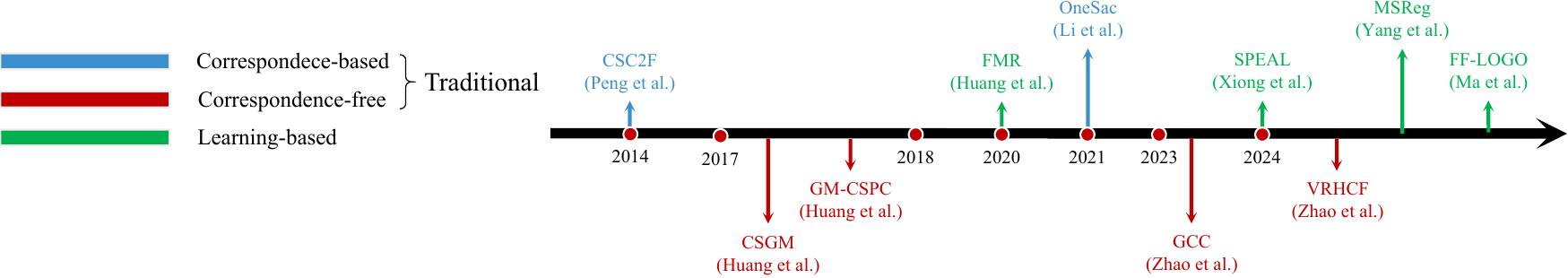}
    \caption{Chronological overview of representative 3D cross-source point cloud registration methods.}
    \label{fig:cross-source axis}
\end{figure*}

Cross-source point cloud registration is vital to merge multi-modal point clouds. The main challenge is the dramatic point distribution variation. This section reviews methodologies in cross-source point cloud registration. The taxonomy, chronological overview, and performance comparison are shown in Fig.~\ref{fig:cross-source}, Fig.~\ref{fig:cross-source axis} and Table~\ref{cross-source}, respectively.


\begin{table*}[t]
    \centering
    \caption{Performance summary of typical cross-source 3D point cloud registration methods.}
    \vspace{-0.2cm}
    \fontsize{4}{5}\selectfont
    \resizebox*{\textwidth}{!}{
\setlength{\arrayrulewidth}{0.2pt}
\begin{tabular}{|c|l|c|c|c|}
\hline
Year & Method          & Data Type                & Category                & Performance                                                             \\ \hline
2014   & CSC2F~\cite{peng2014Street}        & Point cloud              & Correspondence-based    & Robust to density, noise, scale and occlusion differences               \\ \hline
\multirow{2}{*}{2017} & CSGM~\cite{huang2017Systematic}       & Point cloud              & Correspondence-free             & Outperforms CPD                          \\ \cline{2-5} 
                      & GM-CSPC~\cite{huang2017CF}       & Point cloud              & Correspondence-free             & Outperforms CPD                                                 \\ \hline
2020  & FMR~\cite{huang2020FMR}       & Point cloud              & Learning-based & Robust to noise, outliers and density differences                       \\ \hline
2021  & OneSac~\cite{li2021point}          & Point cloud              & Correspondence-based    & Robust against severe outliers                                          \\ \hline
2023  & GCC~\cite{zhao2023Accurate}        & Point cloud              & Correspondence-free    & Outperforms CICP, GCTR, FGR, DGR, FMR                           \\ \hline
\multirow{4}{*}{2024} & SPEAL~\cite{xiong2024Speal}       & Point cloud              & Learning-based        & Outperforms FGR, DGR                          \\ \cline{2-5} 
                      & VRHCF~\cite{zhao2024VRHCF}        & Point cloud              & Correspondence-free    & Outperforms GICP, GCTR, GCC                          \\ \cline{2-5}
                      & MSReg~\cite{yang2024novel}        & Point cloud              & Learning-based    & Effective for large-scale and heterogeneous data                          \\ \cline{2-5}
                      & FF-LOGO~\cite{ma2024FF-LOGO}          & Point cloud              & Learning-based        & Outperforms CICP, GCTR, DGR, FMR, GCC                                   \\ \hline
\end{tabular}}
\label{cross-source}
\end{table*}
\subsubsection{\textbf{Traditional Methods}}
These methods can be further categorized into correspondence-based and correspondence-free.

{\textit{(i)}} \textbf{Correspondence-based methods}.
These methods generally design scale-aware descriptors for correspondence generation followed by a transformation estimator. Persad and Armenakis~\cite{persad2017Automatic} proposed to perform registration in 2D projection images by matching scale, rotation, and translation-invariant descriptors on sparse 2D keypoints. Some methods perform registration directly in the 3D space.  For instance, Peng et al.~\cite{peng2014Street} proposed a two-stage matching process. It matches ensemble of shape functions (ESF) descriptors in the coarse stage, followed by ICP refinement. In a transformation-decomposition way, Li et al.~\cite{li20211PRANSAC} decomposed the full seven-parameter
registration problem into three subproblems, i.e., scale, rotation, and translation, through line vectors.


{\textit{(ii)}} \textbf{Correspondence-free methods}.
These methods register point clouds by leveraging global structures or statistical models. Graph-based approaches, such as~\cite{huang2017Systematic}, apply graph matching to represent macro and micro structures. Statistical modeling and refinement methods~\cite{huang2017CF, zhao2023Accurate, zhao2024VRHCF} focus on probabilistic representations of point clouds, using models such as GMM to capture global structures and iteratively refine alignments. 

\subsubsection{\textbf{Learning-based Methods}}
Learning-based methods handle noise, outliers, and density variations by designing robust feature learning modules. Huang et al~.\cite{huang2020FMR} introduced a feature-metric framework that bypasses correspondence search, accelerating the process while maintaining the robustness to noise. Xiong et al~.\cite{xiong2024Speal} extended this by proposing skeletal representations to enhance topological feature encoding. Building on this, Yang et al.~\cite{yang2024novel} advanced correspondence refinement through a descriptor with in-plane rotation equivalence and disparity-weighted scoring. Recently, Ma et al.~\cite{ma2024FF-LOGO} developed FF-LOGO, integrating feature filtering with local-global optimization to ensure modality-invariant feature extraction. 

Despite recent advances, it is still challenging to handle data with severe point density variation.

\subsection{Color Point Cloud Registration}
\begin{figure}
    \centering
    \includegraphics[width=0.29\textwidth]{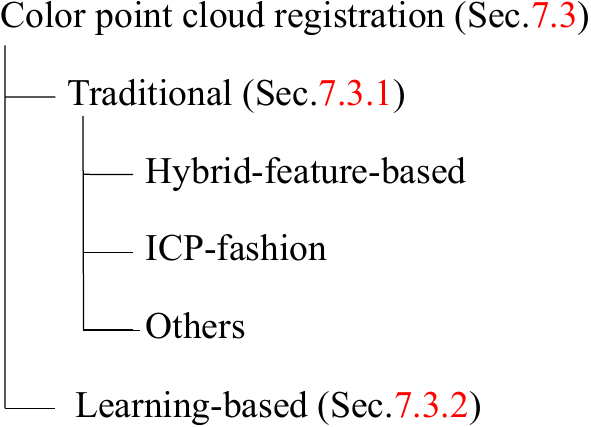}
    \caption{Taxonomy of 3D color point cloud registration methods.}
    \label{fig:color}
    \vspace{-0.45cm}
\end{figure}

\begin{figure*}[t]
    \centering
    \includegraphics[width=1\textwidth]{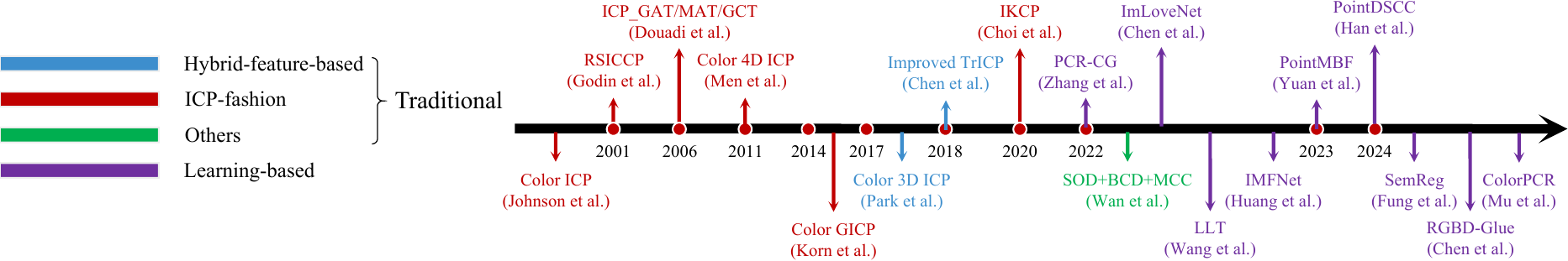}
    \caption{Chronological overview of representative 3D color point cloud registration methods.}
    \label{fig:color axis}
\end{figure*}

For point clouds acquired by RGB-D sensors, additional color cues can be leveraged for point cloud registration. Existing methods can be categorized into traditional and learning-based. The taxonomy, chronological overview, and performance comparison are shown in Fig.~\ref{fig:color}, Fig.~\ref{fig:color axis} and Table~\ref{color}, respectively.


\begin{table*}[t]
    \centering
    \caption{Performance summary of typical 3D color point cloud registration methods.}
    \vspace{-0.2cm}
    \resizebox*{\textwidth}{!}{
\begin{tabular}{|c|l|c|c|c|}
\hline
Year & Method           & Data Type   & Category    & Performance                                                     \\ \hline
1999   & Color 6D ICP~\cite{johnson1999Registration} & Point cloud & ICP-fashion      & Incorporates both 3D data and color information                 \\ \hline
2001   & RSICCP~\cite{godin2001Method}   & Range image & ICP-fashion      & Invariant to rigid transformations                              \\ \hline
2006   & ICP\_GAT/MAT/GCT~\cite{douadi2006Pair}  & Range image & ICP-fashion      & Integrates color information in ICP variants                    \\ \hline
2011   & Color 4D ICP~\cite{men2011color}     & Point cloud & ICP-fashion      & Combines point range and weighted hue values                    \\ \hline
2014   & Color GICP~\cite{korn2014color}    & Point cloud & ICP-fashion      & Robust to low-quality color images                              \\ \hline
2017   & Color 3D ICP~\cite{park2017Colored}    & Point cloud & Hybrid-feature-based        & Outperforms Color 6D ICP, Color 4D ICP, Color GICP              \\ \hline
2018   & Improved TrICP~\cite{chen2018Registration}    & Point cloud & Hybrid-feature-based      & Outperforms TrICP                                               \\ \hline
2020   & IKCP~\cite{choi2020Iterative}    & Point cloud & ICP-fashion      & Outperforms Color 6D ICP, Color 3D ICP, Color GICP              \\ \hline
\multirow{5}{*}{2022} & PCR-CG~\cite{zhang2022PCR-CG}   & Point cloud & Learning-based & Outperforms Color 3D ICP                          \\ \cline{2-5} 
                      & SOD+BCD+MCC~\cite{wan2021RGB}     & Point cloud & Others      & Addresses data degradation and uneven distribution                          \\ \cline{2-5} 
                      & ImLoveNet~\cite{chen2022imlovenet}   & Point cloud, color image & Learning-based & Tackles the challenge of low-confidence features                          \\ \cline{2-5} 
                      & LLT~\cite{wang2022improving}   & Point cloud, RGB-D image & Learning-based & Addresses challenges in multi-modal integration for registration                          \\ \cline{2-5} 
                      & IMFNet~\cite{huang2022IMFNet}   & Point cloud, image & Learning-based & Integrates weighted texture information                                        \\ \hline
2023  & PointMBF~\cite{yuan2023PointMBF}   & Point cloud, RGB-D image & Learning-based & Outperforms LLT                                        \\ \hline
\multirow{4}{*}{2024} & PointDSCC~\cite{han2024Point}     & Point cloud & Learning-based & Outperforms Color 3D ICP, Color GICP                          \\ \cline{2-5} 
                      & SemReg~\cite{fung2025semreg}   & Point cloud, 2D image & Learning-based & Outperforms PCR-CG                          \\ \cline{2-5} 
                      & RGBD-Glue~\cite{chen2024rgbd}   & Point cloud & Learning-based & Outperforms LLT, PointMBF                          \\ \cline{2-5} 
                      & ColorPCR~\cite{mu2024colorpcr}      & Point cloud & Learning-based & Addresses challenges in low-overlap point cloud registration    \\ \hline
\end{tabular}}
\label{color}
\end{table*}
\subsubsection{\textbf{Traditional Methods}}
We further classify them as Hybrid-feature-based, ICP-fashion, and the others.

{\textit{(i)}} \textbf{Hybrid-feature-based methods}.
Hybrid-feature-based methods enhance point cloud registration by combining color, texture, and geometric features to establish correspondences. It can be categorized into two groups: traditional-feature-based and color-enhanced-feature-based. Traditional-feature-based approaches~\cite{ye2018Study, chen2018Registration, wan2019RGB} primarily rely on classical feature extraction techniques and traditional methods for integrating color cues into the registration process. Color-enhanced-feature-based approaches~\cite{park2017Colored, yang2020Color} refine geometric feature matching by incorporating color for enhanced robustness. 

{\textit{(ii)}} \textbf{ICP-fashion methods}.
ICP-fashion methods adapt the ICP method by incorporating color and other features to improve registration accuracy. Early works~\cite{johnson1999Registration, godin2001Method, douadi2006Pair} incorporate invariant attributes and color range images to strengthen matching. Later, Men et al.~\cite{men2011color} and Korn et al.~\cite{korn2014color} combined geometry with hue and $L^{\star}A^{\star}B^{\star}$ color space to accelerate convergence and improve transformation accuracy. More recently, Choi et al.~\cite{choi2020Iterative, choi2021Colored} refined pose and depth using color-supported soft matching, addressing depth errors with adaptive cost functions. 

{\textit{(iii)}} \textbf{Others}.
Recent advances in color point cloud registration focus on improving robustness under challenging conditions. Liu et al.~\cite{liu2022Genetic} used a genetic algorithm to improve the robustness of the registration under changing lighting conditions. Wan et al.~\cite{wan2021RGB} applied salient object detection and bidirectional color distance for precise structural registration. 

\subsubsection{\textbf{Learning-based Methods}}
These methods can be further categorized into three main approaches: 2D-3D fusion, RGB-D fusion and hierarchical fusion. 2D-3D fusion approaches, such as PCR-CG~\cite{zhang2022PCR-CG}, ImLoveNet~\cite{chen2022imlovenet} and SemReg~\cite{fung2025semreg}, integrate color features from 2D images with 3D geometry to enhance registration performance. RGB-D fusion approaches~\cite{wang2022improving, yuan2023PointMBF, han2024Point, zhou2024boosting, chen2024rgbd} focus on combining RGB and depth data to exploit their complementary strengths for better correspondence estimation. Hierarchical fusion approaches, such as IMFNet~\cite{huang2022IMFNet}, ColorPCR~\cite{mu2024colorpcr} and ~\cite{yan2024discriminative}, improve registration by combining multi-level fusion strategies and enhancing feature interpretability. 


\subsection{Multi-instance Point Cloud Registration}
\begin{figure}
    \centering
    \includegraphics[width=0.34\textwidth]{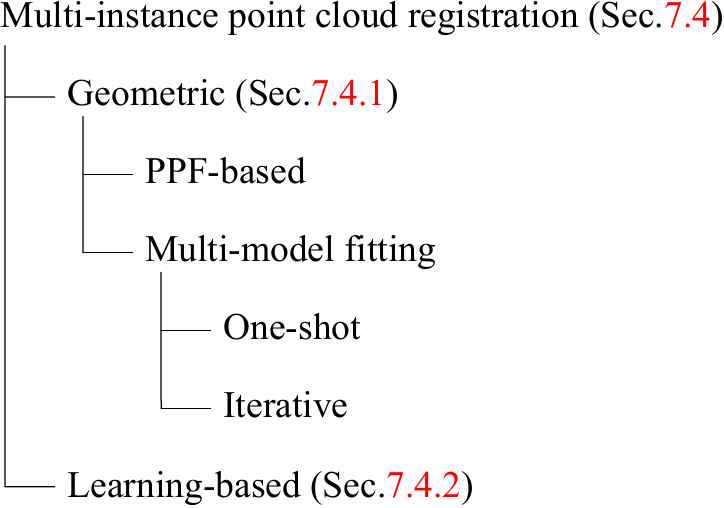}
    \caption{Taxonomy of 3D multi-instance point cloud registration methods.}
    \label{fig:multi-instance}
    \vspace{-0.45cm}
\end{figure}

\begin{figure*}[t]
    \centering
    \includegraphics[width=1\textwidth]{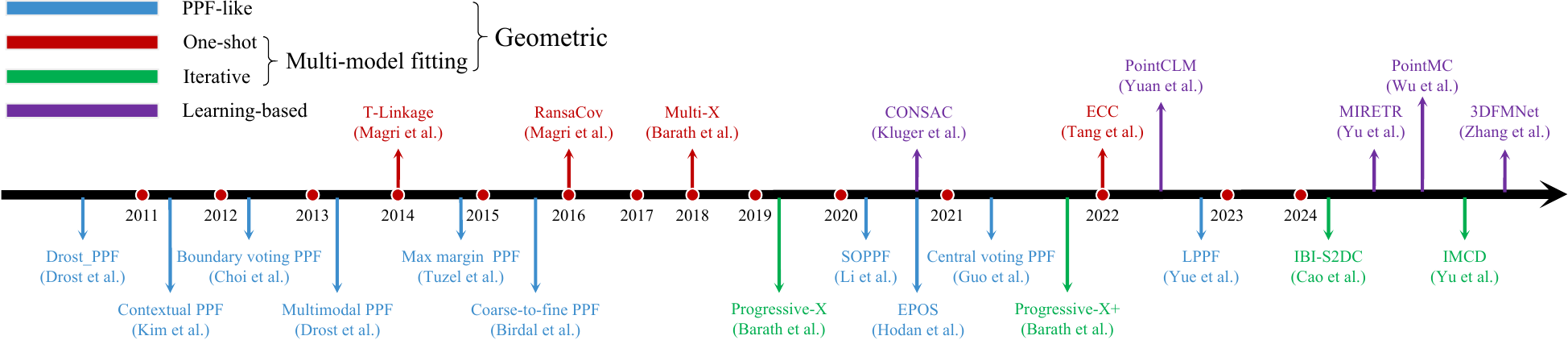}
    \caption{Chronological overview of representative 3D multi-instance point cloud registration methods.}
    \label{fig:multi-instance axis}
\end{figure*}

Multi-instance point cloud registration estimates transformations of multiple instances in the scene. However, this process comes with challenges such as the ambiguity of correspondences from different instances, occlusions, and noise. Research in this area generally falls into two categories: geometric and learning-based. The taxonomy, chronological overview, and performance comparison are shown in Fig.~\ref{fig:multi-instance}, Fig.~\ref{fig:multi-instance axis} and Table~\ref{multi-instance}, respectively.


\begin{table*}[t]
    \centering
    \caption{Performance summary of typical 3D multi-instance point cloud registration methods.}
    \vspace{-0.2cm}
    \resizebox*{\textwidth}{!}{
\begin{tabular}{|c|l|c|c|c|}
\hline
Year & Method          & Data Type                & Category  & Performance                                                           \\ \hline
2010   & Drost\_PPF~\cite{drost2010Drost_PPF}  & Point cloud              & PPF-based    & Robust to noise and clutter                                           \\ \hline
2011   & Contextual PPF~\cite{kim20113D}    & Range image              & PPF-based    & Outperforms Drost\_PPF                                                \\ \hline
\multirow{2}{*}{2012} & Boundary voting PPF~\cite{choi2012Voting}   & CAD model                & PPF-based    & Exploits boundary information, suitable for planar objects                          \\ \cline{2-5} 
                      & Multimodal PPF~\cite{drost20123D}  & Range image              & PPF-based    & Combines intensity and depth data, invariant to scale and rotation    \\ \hline
\multirow{2}{*}{2014} & T-Linkage~\cite{magri2014T-Linkage}  & Point cloud, image pair    & One-shot  & Outperforms J-Linkage                          \\ \cline{2-5} 
                      & Max margin PPF~\cite{tuzel2014Learning}  & Point cloud, 3D model    & PPF-based    & Outperforms Drost\_PPF, suitable for self-similar and planar surfaces \\ \hline
2015   & Coarse-to-fine PPF~\cite{birdal2015PPF} & Point cloud, depth image & PPF-based    & Outperforms Drost\_PPF                                                \\ \hline
2016   & RansaCov~\cite{magri2016RansaCov}  & Image pair               & One-shot  & Outperforms J-Linkage, T-Linkage                                      \\ \hline
2018   & Multi-X~\cite{barath2018Multi-X} & Image pair               & One-shot  & Outperforms T-Linkage                                                 \\ \hline
2019  & Progressive-X~\cite{barath2019Progressive-X} & Point cloud, image pair    & Iterative  & Outperforms T-Linkage, RansaCov. Multi-X                              \\ \hline
\multirow{3}{*}{2020} & SOPPF~\cite{li20203D}     & Point cloud              & PPF-based    & Outperforms Drost\_PPF                          \\ \cline{2-5} 
                      & EPOS~\cite{hodan2020Epos}  & RGB image                & PPF-based    & Outperforms Drost\_PPF                          \\ \cline{2-5} 
                      & CONSAC~\cite{kluger2020CONSAC} & Image pair               & Learning-based  & Outperforms T-Linkage, RansaCov. Multi-X, Progressive-X               \\ \hline
\multirow{2}{*}{2021} & Central voting PPF~\cite{guo2021Central_Voting_PPF}    & Point cloud, image       & PPF-based    & Outperforms Drost\_PPF                          \\ \cline{2-5} 
                      & Progressive-X+~\cite{barath2021Progressive-X+} & Image pair               & Iterative  & Outperforms T-Linkage, RansaCov, Multi-X, Progressive-X, CONSAC       \\ \hline
\multirow{3}{*}{2022} & ECC~\cite{tang2022ECC}   & Point cloud              & One-shot  & Outperforms T-Linkage, Progressive-X, CONSAC                          \\ \cline{2-5} 
                      & PointCLM~\cite{yuan2022PointCLM}   & Point cloud              & Learning-based  & Outperforms T-Linkage, RansaCov, CONSAC, ECC                          \\ \cline{2-5} 
                      & LPPF~\cite{yue2022CF}    & Point cloud              & PPF-based    & Robust to noise, fast convergence                                     \\ \hline
\multirow{5}{*}{2024} & IBI-S2DC~\cite{cao2024IBI}    & Point cloud              & Iterative & Outperforms T-Linkage, Progressive-X, CONSAC, ECC, PointCLM                          \\ \cline{2-5} 
                      & MIRETR~\cite{yu2024MIRETR}     & Point cloud              & Learning-based      & Outperforms T-Linkage, RansaCov, ECC, PointCLM                          \\ \cline{2-5} 
                      & PointMC~\cite{wu2024PointMC}     & Point cloud              & Learning-based & Outperforms T-Linkage, RansaCov, CONSAC, ECC, PointCLM                          \\ \cline{2-5} 
                      & IMCD~\cite{yu2024IMCD}     & Point cloud              & Iterative & Outperforms ECC, PointCLM                          \\ \cline{2-5} 
                      & 3DFMNet~\cite{zhang20243d}  & Point cloud              & Learning-based & Outperforms T-Linkage, RansaCov, ECC, PointCLM, MIRETR                \\ \hline
\end{tabular}}
\label{multi-instance}
\end{table*}
\subsubsection{\textbf{Geometric Methods}}
Geometric methods leverage spatial and structural information to align multiple instances within a scene. These approaches can be categorized into point-pair-features-based (PPF-based) and multi-modal fitting techniques. 

{\textit{(i)}} \textbf{PPF-based methods}.
Early multi-instance registration methods focus on correspondence-free techniques for object recognition and pose estimation. They leverage PPF to encode geometric relationships between points, efficiently estimating multiple poses. Drost et al.~\cite{drost2010Drost_PPF} laid the foundation by introducing a global model description and a fast voting scheme to generate pose hypotheses efficiently. Building on this, many follow-ups have been proposed. Vidal et al.~\cite{vidal2018Method} introduced surface information to improve feature extraction as a preprocessing step. In the modeling stage, various strategies~\cite{kim20113D, choi2012Voting, tuzel2014Learning, drost20123D, li20203D, hodan2020Epos} are developed to optimize feature discrimination. In the voting stage, Guo et al~.\cite{guo2021Central_Voting_PPF} proposed a center voting strategy based on geometric relationships to improve pose hypothesis generation. In addition, Vock et al~.\cite{vock2019Fast} developed an efficient hypothesis validation method to reduce false positives. In a coarse-to-fine fashion, Birdal et al~.\cite{birdal2015PPF} and Yue et al~.\cite{yue2022CF} combined initial segmentation with progressively refined pose estimation.

{\textit{(ii)}} \textbf{Multi-model fitting methods}.
These methods estimate model parameters from data points generated by multiple models. They can be broadly classified into two categories: one-shot and iterative.

{\textit{1)}} \textbf{One-shot manner}. 
These methods estimate model parameters in a single step for efficient fitting. 
Clustering-based approaches, such as T-Linkage~\cite{magri2014T-Linkage}, focus on grouping correspondences to detect model instances. Magri and Fusiello ~\cite{magri2016RansaCov} developed RansaCov, formulating the problem as set coverage, effectively managing intersecting structures and outliers. Optimization-based methods, such as Multi-X~\cite{barath2018Multi-X}, employ optimization techniques for multi-class and multi-instance fitting.
Some other methods direct clustering transformation hypotheses. An example is ECC~\cite{tang2022ECC}, which directly groups noisy correspondences via compatibility matrices.

{\textit{2)}} \textbf{Iterative manner}. 
Iterative methods successively register instances. Baráth et al~.\cite{barath2019Progressive-X, barath2021Progressive-X+} iteratively improved multi-model parameter estimation by combining hypothesis generation with optimization techniques. Cao et al.~\cite{cao2024IBI} introduced the first iterative framework IBI specifically designed for 3D multi-instance registration, which mines consistent seed correspondences to guide inlier recovery. Later, Yu et al.~\cite{yu2024IMCD} enhanced clustering efficiency and robustness without known cluster number under an iterative paradigm.

\subsubsection{\textbf{Learning-based Methods}}
Learning-based methods have progressed to address challenges such as clutter, occlusion, and overlapping instances. Some methods learn correspondences from data or learn multiple registration poses from correspondences. For instance, CONSAC~\cite{kluger2020CONSAC} learns sampling weights to guide RANSAC, and PointMC~\cite{wu2024PointMC} learns correspondences to performance pose clustering. In addition, PointCLM~\cite{yuan2022PointCLM} estimates multi-instance registration poses from correspondences with contrastive learning. A recent trend is end-to-end learning. 
MIRETR~\cite{yu2024MIRETR} proposes a coarse-to-fine transformer-based approach to extract instance-aware correspondences and predict registration poses. Recently, 3DFMNet~\cite{zhang20243d} introduces a novel framework that first focuses on object proposals and then performs instance-level registration.

\subsection{Summary}
The following points can be summarized.

1) {\textbf {Cross-scale registration.}} Geometric methods are still the mainstream approaches. Simultaneously addressing scale and partial overlap issues remains a challenge.

2) {\textbf {Cross-source registration.}} Recent deep-learning-based methods show powerful ability. However, it is still challenging to align cross-source point clouds with significant point distribution variation.

3) \textbf{Color point cloud registration.} A recent trend is to guide geometric registration with color cues for either performance boosting or supervision signal reduction.

4) \textbf{Multi-instance registration.} End-to-end learning methods have shown better performance than methods learning with correspondences only. Both bottom-up and top-down methods have shown their own merits.

\section{Challenges and Opportunities}\label{sec:sum}
Although we have witnessed great success towards robust 3D point cloud registration in the last decades, there are still several critical issues requiring future research attention. 
\begin{enumerate}
    \item \textbf{Push unsupervised registration to the limit.} The point cloud registration problem, in its nature, is an optimization problem. Although fully-supervised methods have achieved great success in recent years, unsupervised methods are more applicable in real-world applications and are very promising when combined with geometric constraints. We have witnessed a few unsupervised methods at present; their performance is even comparable with supervised methods, motivating us to explore the performance upper bound.
    \item \textbf{End-to-end learning or hybrid solutions.} End-to-end learning is a well-known fashion to tackle 2D vision problems. Following this trend, a number of end-to-end registration frameworks emerge. However, a recent work demonstrates that a decoupled framework (e.g., learned features with geometrical estimators) surpasses state-of-the-art methods, and enables better generalization ability. As such, this problem should be investigated more deeply to give an answer to the 3D registration community.
     \item \textbf{Robust transformation estimation with extremely scare inliers.} Correspondence-based registration methods are frequently revisited due to their promising performance and robustness. Since the invention of RANSAC, many follow-ups have been proposed. However, the problem is extremely ill-posed in the presence of scares inliers, and existing methods still fail to achieve robust results. New datasets and experimental settings toward this direction deserve more attention.
     \item \textbf{Ultra-small residual registration error elimination.} For fine registration, ICP and its variants are arguably de facto choices. An interesting observation is that most fine registration works tend to tackle the problem of global registration, while current state-of-the-art coarse registration methods could already present a good initial guess. In many industrial and measuring applications, ultra-small residual error elimination in the presence of noise, weak–geometric features is of great demand. However, existing fine registration methods struggle to deal with such situation.
     \item \textbf{Multi-view registration in the wild.} Current multi-view registration methods assume that the object/scene-of-interest are static, the scanning overlap are ensured, and the scanned data are controlled. Performing multi-view registration with point cloud sequences captured from a dynamic and unknown scene, remains an open problem.
     \item \textbf{The registration of 3D Gaussians.} 3D Gaussian representations open a new era for rendering and 3D reconstruction. In large scene rendering problems, a few trails on registering 3D Gaussians have been made. We believe there are many interesting problems to be explored, given mature point cloud registration solutions and new 3D representations.
     \item \textbf{Cross-scale 3D registration is challenging.} The scale factor sometimes is ambiguous in applications such as autonomous navigation and robotics, the problem is even more challenging when other nuisances exist simultaneously. However, we find that most of the research efforts have been paid compared to the standard registration problem without scale variation. This direction is more challenging.
     \item \textbf{Registration with large pretrained models.} Large pretrained models have advanced the multi-modal generation tasks greatly. The strong generation ability of large pretrained model enables shape completion. This is supposed to improve the registration performance for low-and-non-overlapping data. 
\end{enumerate}\par
\section{Conclusions}\label{sec:conc}
This paper provides a comprehensive overview of 3D point cloud registration methods in the last three decades, covering a broad area of registration problems.
We have presented a comprehensive taxonomy and
performance comparison of reviewed methods. The traits,
merits, and demerits of these methods have been summarized.
Finally, several future research directions have been discussed.
\section*{Acknowledgments}
The authors would like thank Dr. Siwen Quan for the discussion during the work.
\bibliographystyle{IEEEtran}
\bibliography{mybibfile}

\end{document}